\documentclass[11pt, a4paper]{article}
\usepackage[utf8]{inputenc}
\usepackage[T1]{fontenc}
\usepackage{lineno}
\usepackage{graphicx}
\usepackage{geometry}
\usepackage{array}
\usepackage{booktabs}
\usepackage{textcomp}
\usepackage{algorithm}
\usepackage{algpseudocode}
\usepackage{phaistos}
\usepackage{graphicx}
\usepackage{subcaption}
\usepackage{caption}
\usepackage{xtab}
\usepackage{multicol}
\usepackage{tabularx}
\usepackage{longtable}
\usepackage{adjustbox}
\usepackage{multirow}
\usepackage{svg}
\usepackage{amssymb}
\usepackage{amsmath}
\usepackage{xspace}
\usepackage{authblk}
\usepackage{dsfont}
\usepackage{lineno}

\usepackage[numbers,compress]{natbib}

\usepackage{hyperref}
\hypersetup{
    colorlinks=true,
    linkcolor=blue,
    filecolor=magenta,      
    urlcolor=cyan,
    pdftitle={Overleaf Example},
    pdfpagemode=FullScreen,
    }
\date{}
\DeclareUnicodeCharacter{2032}{\ensuremath{'}}
\DeclareUnicodeCharacter{00B0}{\ensuremath{^\circ}} 
\DeclareUnicodeCharacter{2212}{-}

\DeclareMathOperator{\Dead}{{\textit{Dead coral}}}

\DeclareMathOperator{\AcroporeB}{\textit{Acropora Branching}}
\DeclareMathOperator{\AcroporeD}{\textit{Acropora Digitate}}

\DeclareMathOperator{\AcroporeT}{\textit{Acropora Tabular}}

\DeclareMathOperator{\NoAcroporeE}{\textit{Non-acropora Encrusting}}

\DeclareMathOperator{\NoAcroporeM}{\textit{Non-acropora Massive}}
\DeclareMathOperator{\NoAcroporeSu}{\textit{Non-acropora Submassive}}

\DeclareMathOperator{\Millepore}{\textit{Non-acropora Millepora}}

\DeclareMathOperator{\AlgaeL}{\textit{Algae Coralline}}
\DeclareMathOperator{\AlgaeD}{\textit{Algae Halimeda}}
\DeclareMathOperator{\AlgaeS}{\textit{Algae Turf}}
\DeclareMathOperator{\AlgaeA}{\textit{Algal Assemblage}}
\DeclareMathOperator{\Algae}{\textit{Algae}}
\DeclareMathOperator{\Fish}{{\textit{Fish}}}

\DeclareMathOperator{\SeaU}{\textit{Sea urchin}}
\DeclareMathOperator{\SeaC}{{\textit{Sea cucumber}}}

\DeclareMathOperator{\Homo}{{\textit{Homo Sapiens}}}
\DeclareMathOperator{\Blurred}{{\textit{Blurred}}}

\DeclareMathOperator{\Sand}{\textit{Sand}}
\DeclareMathOperator{\Rubble}{\textit{Rubble}}

\DeclareMathOperator{\Rock}{\textit{Rock}}

\begin{document}
\let\WriteBookmarks\relax
\def\floatpagepagefraction{1}
\def\textpagefraction{.001}

\title{From underwater to aerial: a novel multi-scale knowledge distillation approach for coral reef monitoring} 
\author[1,2, *]{Matteo Contini}
\author[1]{Victor Illien}
\author[4]{Julien Barde}
\author[4]{Sylvain Poulain}
\author[3]{Serge Bernard}
\author[2]{Alexis Joly}
\author[1]{Sylvain Bonhommeau}

\affil[1]{IFREMER Délégation Océan Indien (DOI), Le Port, 97420, La Réunion, France, Rue Jean Bertho}
\affil[2]{INRIA, LIRMM, Université de Montpellier, CNRS, Montpellier, 34000, France}
\affil[3]{CNRS, LIRMM, Université de Montpellier, Montpellier, 34000, France}
\affil[4]{UMR Marbec, IRD, Université de Montpellier, CNRS, Ifremer, Montpellier, 34000, France}

\affil[*]{corresponding author: Matteo Contini (firstname.lastname at ifremer.fr)}

\begin{abstract}
Drone-based remote sensing combined with AI-driven methodologies has shown great potential for accurate mapping and monitoring of coral reef ecosystems.
This study presents a novel multi-scale approach to coral reef monitoring, integrating fine-scale underwater imagery with medium-scale aerial imagery. 
Underwater images are captured using an Autonomous Surface Vehicle (ASV), while aerial images are acquired with an aerial drone.
A transformer-based deep-learning model is trained on underwater images to detect the presence of 31 classes covering various coral morphotypes, associated fauna, and habitats. 
These predictions serve as annotations for training a second model applied to aerial images.
The transfer of information across scales is achieved through a weighted footprint method that accounts for partial overlaps between underwater image footprints and aerial image tiles.
The results show that the multi-scale methodology successfully extends fine-scale classification to larger reef areas, achieving a high degree of accuracy in predicting coral morphotypes and associated habitats.
The method showed a strong alignment between underwater-derived annotations and ground truth data, reflected by an AUC (Area Under the Curve) score of 0.9251.
This shows that the integration of underwater and aerial imagery, supported by deep-learning models, can facilitate scalable and accurate reef assessments.
This study demonstrates the potential of combining multi-scale imaging and AI to facilitate the monitoring and conservation of coral reefs. 
Our approach leverages the strengths of underwater and aerial imagery, ensuring the precision of fine-scale analysis while extending it to cover a broader reef area.

\vspace{0.5cm}


\end{abstract}

\maketitle

\section{Introduction} 
\label{sec_introduction}

Coral reefs are among the richest ecosystems on Earth in terms of species diversity.
Moreover, they provide a number of key services: they function as natural barriers safeguarding coastlines from erosion and extreme weather events and serve as habitats and breeding grounds for innumerable marine species \cite{Hoegh-Guldberg2}.
Additionally, they support local economies by offering resources for fishing, tourism and potential medicinal compounds \cite{rogers}, \cite{bruckner2002life}.
However, these ecosystems are under serious threat from human activities.
Destructive and illegal fishing practices \cite{hidayati2022importance}, anthropogenically derived chemical pollutants \cite{van2011chemical} and coastal development \cite{hughes2003climate} are some of the main causes of coral reef degradation.
Climate change poses an even greater risk through ocean acidification and warming, leading to widespread coral bleaching and habitat loss \cite{hughes2017global}.

In December 2022 the Global Biodiversity Framework was adopted at the 15th Conference of the Parties (COP15) with the objective of protecting 30\% of Earth’s lands, oceans, coastal areas, and inland water by 2030 \cite{noauthor_cop15_nodate}.
This ambitious goal requires the development of innovative monitoring techniques to assess the status of marine ecosystems and guide conservation efforts.

New techniques based on deep learning are emerging, offering the potential to revolutionize marine monitoring.
In \cite{lamperti2023new, https://doi.org/10.1111/2041-210X.14430} the authors developed a deep-learning based method for monitoring marine biodiversity using environmental DNA (eDNA). 
Besides, \cite{Morand2023} proposed the use of convolutional neural networks (CNNs) to predict species distributions in the open ocean by leveraging environmental data and species occurrences.
Artificial Intelligence (AI) models can also be used, coupled with Autonomous Surface Vehicles (ASVs), to classify spatialised underwater images generating high resolution species distribution maps \cite{MISIUK2024108599}
This approach enables fine-grained annotations and predictions, allowing the distinction of coral morphotypes and the identification of specific classes, such as algae, that are often difficult to discern. 
The main drawback of survey methods based solely on ASVs is that, since underwater images are taken at a very fine scale, it is difficult to cover large areas of coral reefs.
This implies that, when areas on the order of tens of hectares are to be monitored, the method poses challenges in terms of cost, processing time and ease of deployment.

Recent advances in imaging technologies have opened up new possibilities for large-scale coral reef monitoring. 
Drone-based imaging has emerged as a valuable tool for coastal habitat mapping and monitoring, providing a cost-effective method for high-resolution habitat classification when combined with machine learning techniques \cite{DOUKARI2022100726}.
The main limitation of using only aerial images is that they do not directly provide detailed information on individual benthic organisms. 
Thus, the annotation of aerial images is usually made on broader classes (e.g., hard bottom, mixed substrate, soft bottom and seagrass). 
Here we argue that they contain sufficient information to infer the benthic community as a whole if they are combined with fine-grained annotations inferred from underwater images.

\begin{figure}[ht]
    \centering
    \begin{subfigure}[b]{0.42\textwidth}
        \includegraphics[width=\textwidth]{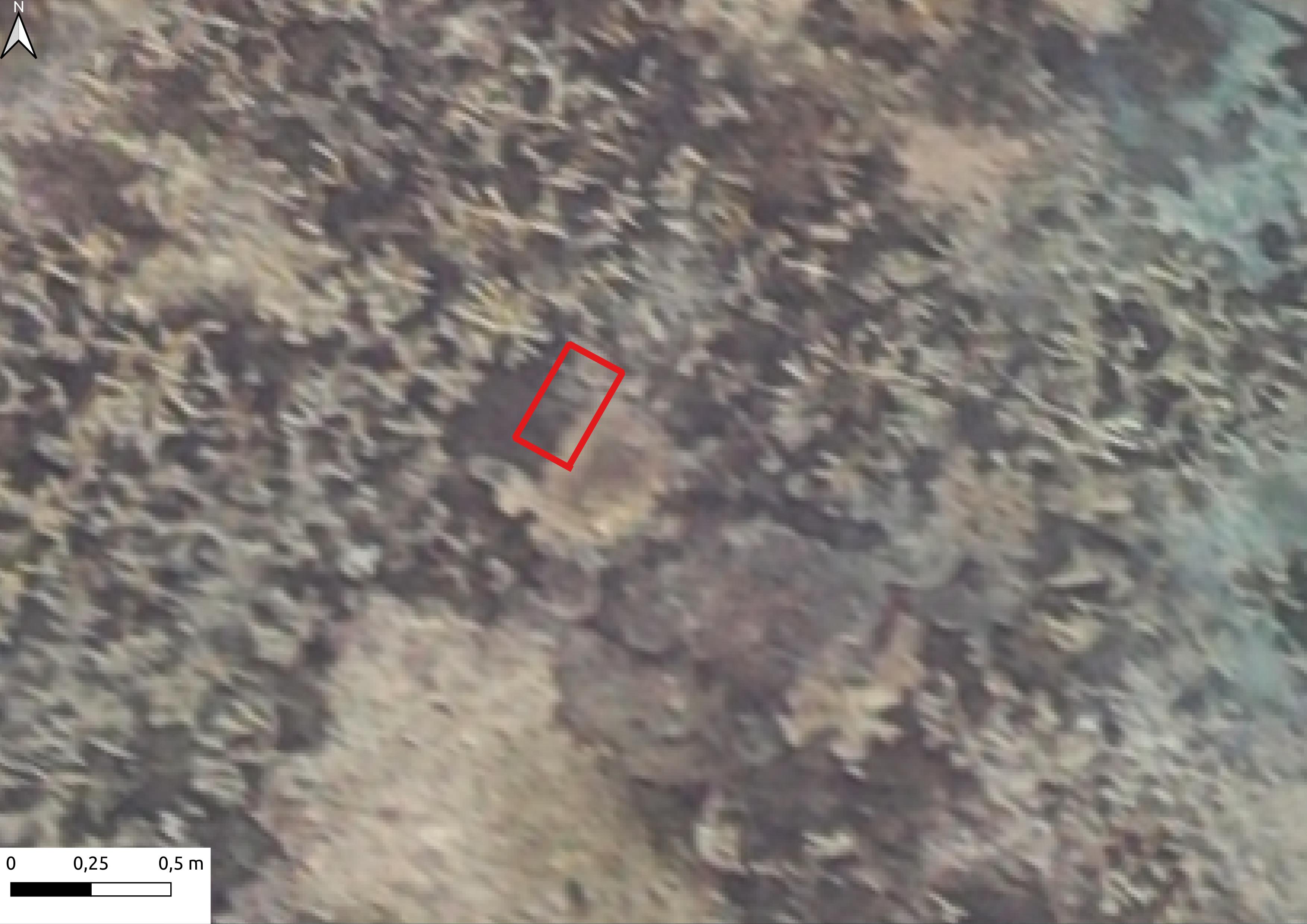}
    \end{subfigure}
    \hspace{2cm}
    \begin{subfigure}[b]{0.42\textwidth}
        \includegraphics[width=\textwidth]{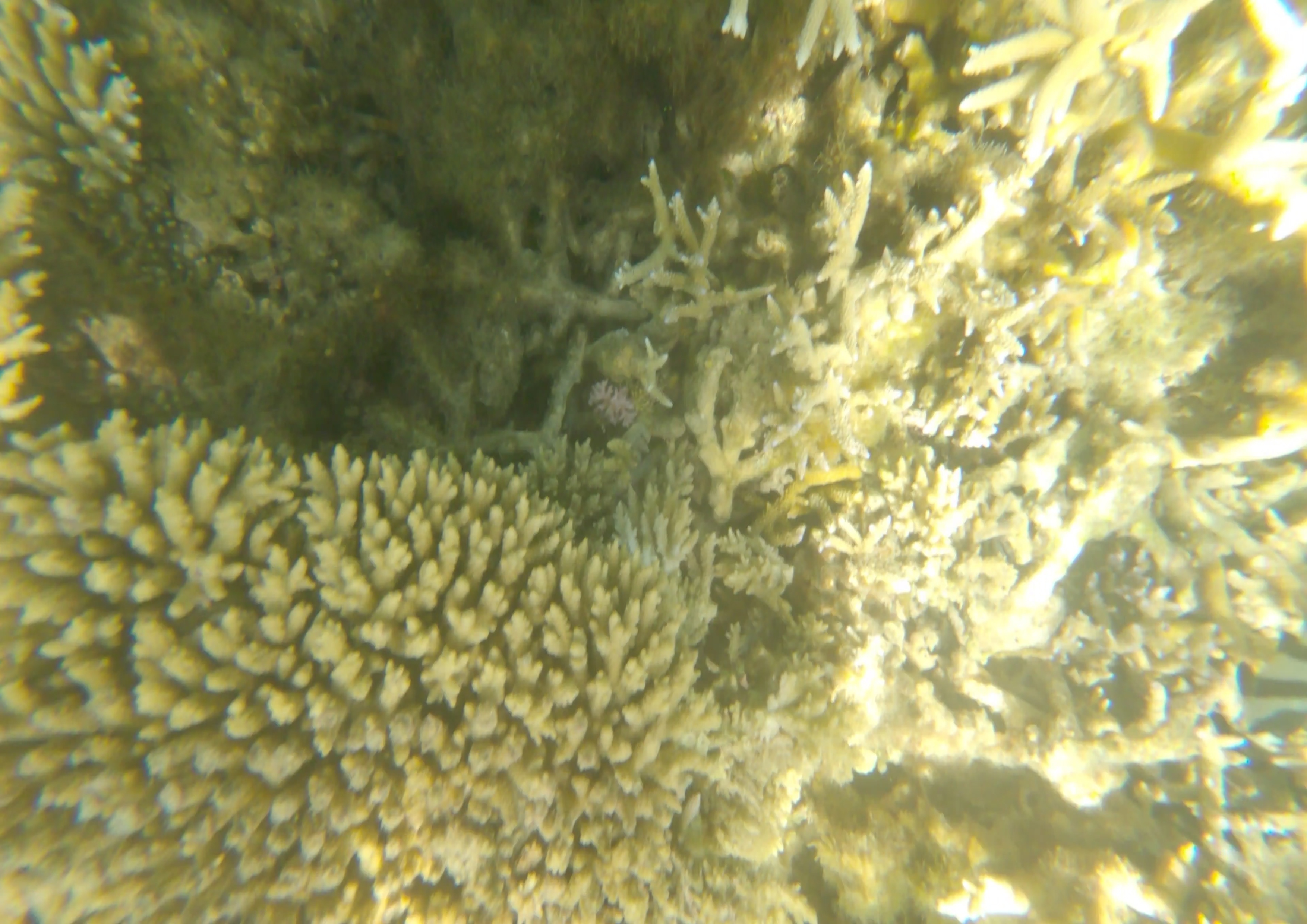}
    \end{subfigure}
    \caption{On the left, an aerial orthophoto captures a complex coral assemblage. 
    However, distinguishing between specific morphotypes is challenging because of the limited resolution. 
    The red rectangle highlights the location of the fine-scale underwater image shown on the right. 
    The underwater image provides a higher level of detail, allowing the identification of distinct coral morphotypes and specific classes, such as $\Algae$, that are difficult to identify in aerial imagery.}
    \label{fig:fine_grained_annotations}
\end{figure}

To give an illustration of this problem,  Figure \ref{fig:fine_grained_annotations} shows an orthophoto obtained from aerial images on the left and an underwater image on the right, collected by an ASV, corresponding to the zone delimited by the red rectangle.
The medium-scale image on the left shows a complex assemblage of corals. 
However, due to the resolution of the image, although one can guess the presence of several morphotypes of corals, identifying them individually remains a difficult task.
This challenge becomes significantly easier when the corresponding underwater images are available. 
The underwater image on the right provides clear distinctions between coral assemblages and other classes, such as $\Algae$, which are often difficult to observe in aerial imagery.
Propagating those fine-grained annotations to the aerial image yields a finer classification of the benthic habitat.

Integrating aerial imagery with human-collected ground truth data can be a first solution to map coastal habitats with high accuracy, as demonstrated in \cite{kvile2024drone}. 
The authors provide a detailed protocol, from drone imagery collection to orthophoto annotation through GIS softwares, allowing the training of segmentation CNN models on aerial images.
In \cite{ventura2023coastal}, the authors use both aerial and underwater imagery, processed with Support Vector Machines (SVM) and Object-Based Image Analysis (OBIA) for benthic habitat classification. 
They used UAVs (Unmanned Aerial Vehicles) to capture high-resolution aerial imagery of coastal areas and USVs to collect ultra-high-resolution underwater images. 
The data from both platforms were processed separately using Structure from Motion (SfM) to create orthophoto mosaics and Digital Surface Models (DSMs). 
These products were then classified using OBIA and SVM algorithms. 
However, their approach does not fully leverage the complementarity of the two data sources. 
Indeed, there is an opportunity to exploit areas where both types of data (fine-scale underwater and medium-scale aerial) are available, in order to improve the model. 
In particular, they do not employ data from the same zones to train models that can infer fine-scale details from medium-scale data in regions where only aerial imagery is available. 
This limitation precludes the potential for synergistic use of overlapping datasets to enhance benthic habitat classification on a broader scale.

The aim of this study is to introduce a novel multi-scale deep-learning approach that integrates underwater and aerial imagery for fine-grained assessment of coral reefs at broad spatial scales.

\begin{figure*}[ht]
    \centering
    \includegraphics[width=\textwidth]{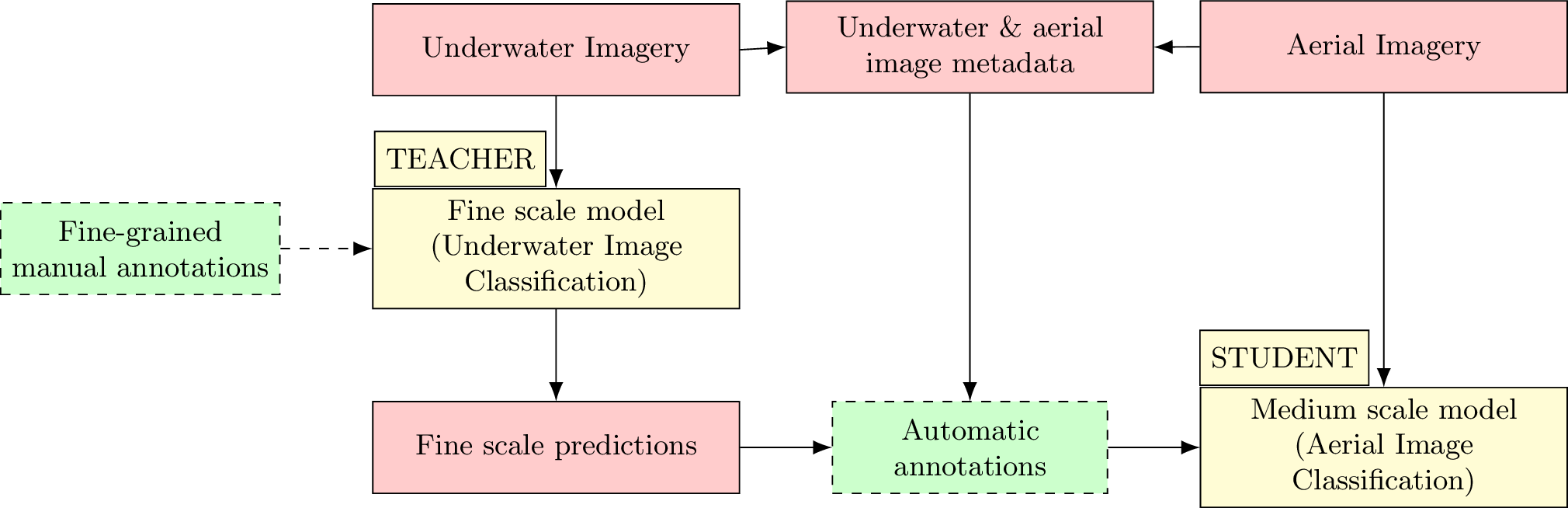}
    \caption{Workflow of the multi-scale approach for coral reef monitoring.}
    \label{fig:workflow}
\end{figure*}

Specifically, we used a transformer-based deep learning model to predict the presence/absence of 31 different classes of corals, associated fauna and habitats in the underwater images. 
To extend this classification to aerial images, we employed the concept of knowledge distillation \cite{gou2021knowledge}, where the underwater model acts as the teacher and the aerial model as the student. 
The objective is for the student model to learn from teacher's outputs, allowing it to achieve comparable performance by mimicking the teacher’s knowledge. 
Concretely, as illustrated in Figure \ref{fig:workflow}, a first fine-scale model (the teacher) is trained on the underwater images associated with fine-grained manual annotations. 
A second model is then trained on the aerial images, using the underwater predictions and image metadata to generate annotations.

The transfer of information across the two scales is achieved through a weighted footprint method that accounts for partial overlaps between underwater image footprints and aerial image tiles.

By reducing the time-consuming annotation process to a single step on underwater images, this approach allows the aerial model to classify images at a larger scale while maintaining as much as possible the fine-scale information provided by the underwater model.

This research offers a powerful tool for coral reef monitoring through an accurate classification of coral morphotypes and associated marine organisms.
Compared to traditional ASV surveying techniques, this method provides significant advantages in terms of cost efficiency, as it reduces the human time required for ASV deployments to cover the same surface area. 
UAVs are also easy to deploy; from the beach, we can access sites that are kilometers away without losing control of the drone. 
Additionally, this approach eliminates the risk of sending ASVs into dangerous zones where they could become stuck or damaged, ensuring safer operations and minimal disturbance to the reef ecosystem.
By demonstrating the combination of advanced technologies such as ASVs, drones, and deep learning models, the study contributes to the development of more effective and efficient conservation strategies, showcasing the potential of multi-scale monitoring for environmental protection.
To our knowledge, this is the first time that such a cascade of deep learning models has been used to classify aerial images.
This opens up new possibilities for upscaling a computer vision model trained on fine-scale images to a larger scale. 
Although being developed for marine applications, this method could be used for terrestrial ecosystems. 

\section{Materials and methods} 
\label{sec_matandmet}

\subsection{Underwater image acquisition}
\label{subsec_underwater_image_acquisition}
Underwater images were collected using an Autonomous Surface Vehicle (ASV) equipped with a \textit{GoPro Hero 8} camera and a differential GPS \textit{Emlid} Reach M2 mounted on a waterproof case. 
The version of the ASV builds on a previous version developed in \cite{gogendeau2024opensourceautonomoussurfacevehicle}.
In order to end up with georeferenced images embedded with attitude metadata (roll, pitch and yaw angles), the following steps were taken:

\begin{enumerate}
    \item Time synchronization between the camera and the GPS clocks.
    \item GPS position correction.
    \item Bathymetry data correction using local geoid parameters and attitude data of the ASV.
    \item Image georeferencing using the corrected GPS position and attitude data.
\end{enumerate}

50 missions were carried out in the lagoon of Reunion Island: 30 in the Saint-Leu lagoon and 20 in the
Trou d’eau lagoon.
For additional details regarding the processes of time synchronization, metadata correction, and other technical aspects, please refer to Appendix \ref{annex:asv} 
where the corresponding subsections are discussed in depth.
For further details on time synchronization, GPS position correction and image georeferencing, please refer to Appendix \ref{annex:asv}, where these aspects are explained in detail.

\subsection{Aerial image acquisition}
\label{subsec_aerial_image_acquisition}

Aerial drone images were taken with a \texttt{DJI Mavic 2 Pro} drone. 
Images were collected following good practices in use in aerial imagery \cite{slocum2019guidelines}. 

Since this drone is not equipped with a differential GPS, once images were taken and the SfM model was built, the orthophoto was georeferenced by collecting ground control points (GCPs) using a differential GPS.

To obtain a high-resolution orthophoto with high positioning precision, the following steps were taken:

\begin{enumerate}
    \item Mission planning: check the equipment, request authorizations from French authorities, weather conditions and plan the flight mission. 
    \item Mission execution: fly the drone at an altitude of 60m  over the area of interest adapting camera settings to the specific conditions of the day. 
    \item Image processing: build the Structure from Motion (SfM) model using images taken during the flight mission. This was done using \texttt{OpenDroneMap}. 
    \item GCP collection: collect GCPs using a GPS with centimetric accuracy. 
    \item Orthophoto georeferencing: georeference the orthophoto using the GCPs. 
\end{enumerate}

Two missions were carried out in the lagoon of Reunion Island: one in the \textit{Saint-Leu} lagoon and the other in the  \textit{Trou d'eau} lagoon.
For further details on mission planning, execution, image processing, and orthophoto georeferencing, please refer to Appendix \ref{annex:uav}, where these aspects are explained in detail.

\subsection{Multi-scale positioning}
\label{subsec_multi_scale_positioning}
Since the objective is to pass information from a fine scale (underwater images) to a larger scale (drone images), the precision of the relative position between underwater and drone images is crucial. 

In order to validate the georeferencing of underwater images with respect to aerial images, we used data from IGN (Institut national de l’information géographique et forestière\footnote{IGN is the French public state administrative establishment that has the main objective of producing and maintaining geographical information for France and its overseas departments and territories}).
Each 3 to 4 years, IGN produces \textit{BD ORTHO®}: a collection of orthophotos with a default resolution of 20 cm. 
The last orthophoto produced by IGN on Reunion island was in 2022, so we decided to use this data as a reference to validate the georeferencing of our data.

\begin{figure}[ht]
    \centering
    \begin{subfigure}[t]{0.4\textwidth}
        \includegraphics[width=\textwidth]{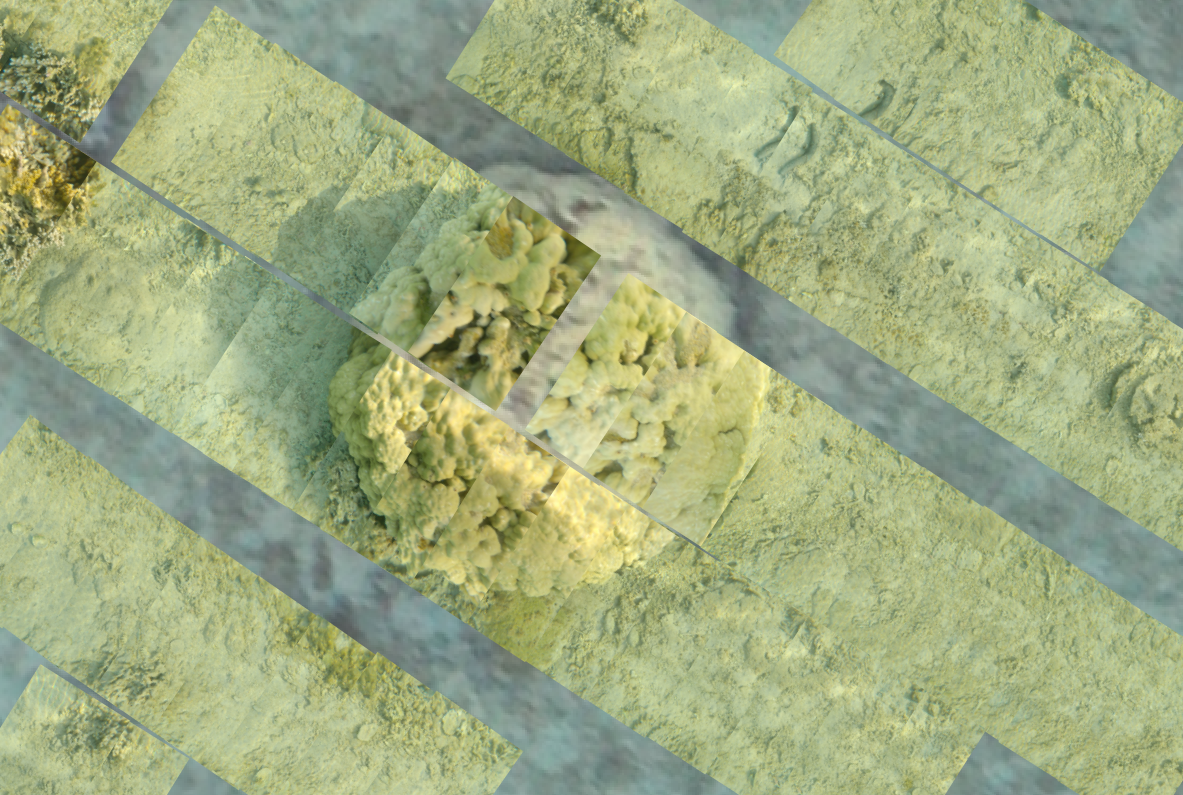}
        \caption{}
        \label{fig:underwater_georeferencing}
    \end{subfigure}
    \hspace{2cm}
    \begin{subfigure}[t]{0.4\textwidth}
        \includegraphics[width=\textwidth]{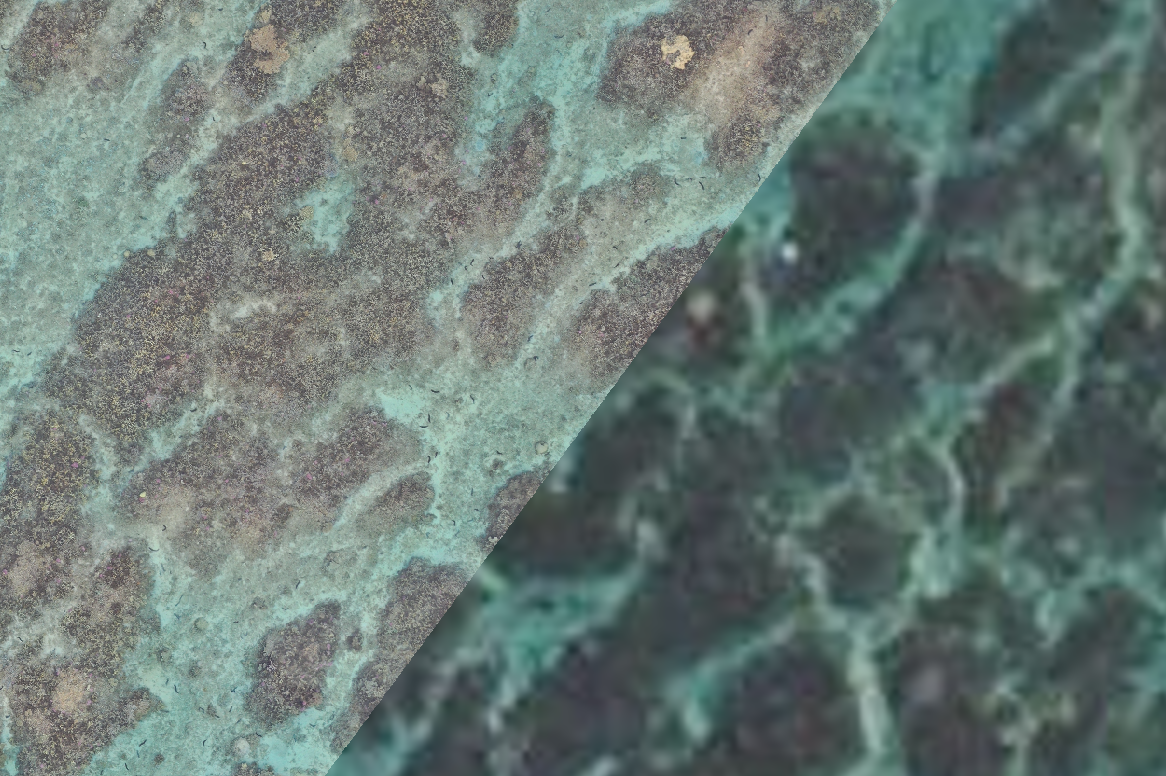}
        \caption{}
        \label{fig:aerial_georeferencing}
    \end{subfigure}
    \caption{Visual georeferencing criteria to validate the georeferencing of underwater and aerial images with respect to the \textit{BD ORTHO®} orthophoto.
    On the left (a) underwater images georeferenced with respect to the aerial orthophoto. 
    On the right (b) aerial orthophoto georeferenced with respect to the \textit{BD ORTHO®} orthophoto produced by the French National Geographic institute (IGN).
    The lighter part on the left corresponds to the drone-based orthophoto and the darker part on the right corresponds to the \textit{BD ORTHO®}.}
    \label{fig:georeferencing}
\end{figure}

Two visual criteria were then used to confirm the georeferencing of the underwater and aerial images with respect to the \textit{BD ORTHO®} orthophoto:

\begin{itemize}
    \item Relative georeferencing: find the presence of easily recognizable objects in both underwater and aerial orthophoto and compare them on a GIS software.
    Often the presence of \textit{Porites} corals can be used to compare the two scales since their contours are easily recognizable in both types of images. 
    See Figure \ref{fig:underwater_georeferencing}.

    \item Aerial absolute georeferencing: find the presence of easily recognizable coral colonies in both aerial and \textit{BD ORTHO®} orthophoto and compare them on a GIS software.
    See Figure \ref{fig:aerial_georeferencing}.
\end{itemize}   

    The combination of the relative georeferencing between underwater and aerial images and the aerial absolute georeferencing with respect to the \textit{BD ORTHO®} orthophoto allowed us to crosscheck the georeferencing of the underwater images with respect to national baseline data.

\subsection{Underwater image classification} 
\label{subsec_underwater_image_classification}
The underwater deep learning model builds on the \textit{DinoV2} architecture, which is a vision transformer model that has been shown to outperform convolutional neural networks on image classification tasks \cite{oquab_dinov2_2023}.
The model has been trained on the open source dataset \href{https://zenodo.org/records/12819157}{Seatizen Atlas image dataset} composed of 51 distinct classes of corals, associated fauna, and habitats.
The model architecture and hyperparameters settings are described in Appendix \ref{sec_finescal_deepmodel}.

Once the model was trained, we ran inference on 56,653 georeferenced images in  \textit{Trou d'eau} lagoon and 58,076 images in \textit{Saint-Leu} lagoon in Reunion Island which are included in the area covered by an aerial drone.

For more information on the data splitting technique used to train the model, please refer to Appendix \ref{sec_data_splitting}.

\subsection{Upscaling predictions} 
\label{subsubsec_upscaling_predictions}

Once the orthophoto is georeferenced and underwater inference is done, the key step is to correctly pass the information from the underwater model to the aerial model. 
The objective is to train an aerial model based on underwater predictions, without spending time on manual annotations of aerial images.

This is achieved by following the steps below:

\begin{enumerate}
    \item Split aerial orthophoto into tiles, ensuring consistency in ground surface representation of each tile across different sessions.
    Each tile represents an area of 1.5m x 1.5m.
    See Section \ref{subsubsec_orthophoto_tiling}.
    \item Filter useless aerial tiles (e.g., black tiles issued from SfM processing errors or tiles without corresponding underwater images).
    See Section \ref{subsubsec_useless_tile_filtering}.
    \item Associate each aerial tile with underwater images whose camera GPS position is within the tile boundaries. 
    See Section \ref{subsubsec_footprint}.
    \item Compute the footprint of each underwater image and filter aerial tiles with not enough underwater coverage.
    See Section \ref{subsubsec_footprint}.
    \item Transform underwater predictions into aerial annotations.
    See Section \ref{subsubsec_transforming_predictions}.
\end{enumerate}

\subsubsection{Orthophoto tiling} 
\label{subsubsec_orthophoto_tiling}

The first step is to split the aerial orthophoto into tiles. 
This is done by taking into account the ground sample distance (GSD) of each orthophoto. 
This approach guarantees that while tile dimensions in pixels may vary due to different GSDs, each tile consistently represents a fixed area on the ground. 
This method allows for standardized comparison and analysis of images across different datasets and sessions, maintaining a consistent spatial resolution. 
Splitting the orthophoto into too small tiles results in images without enough context to be correctly classified and/or with an insufficient resolution. 
On the contrary, splitting the orthophoto into too large tiles results in good classification performances but does not allow for a fine-grained analysis of the data. 
Searching for the best compromise, we fix this area to be 1.5 m x 1.5 m.

\subsubsection{Useless tiles filtering} 
\label{subsubsec_useless_tile_filtering}

\begin{figure}[ht]
    \centering
    \begin{subfigure}[t]{0.28\textwidth}
        \includegraphics[width=\textwidth]{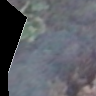}
        \caption{}
        \label{fig:black_tile}
    \end{subfigure}
    \hspace{2cm}
    \begin{subfigure}[t]{0.40\textwidth}
        \includegraphics[width=\textwidth]{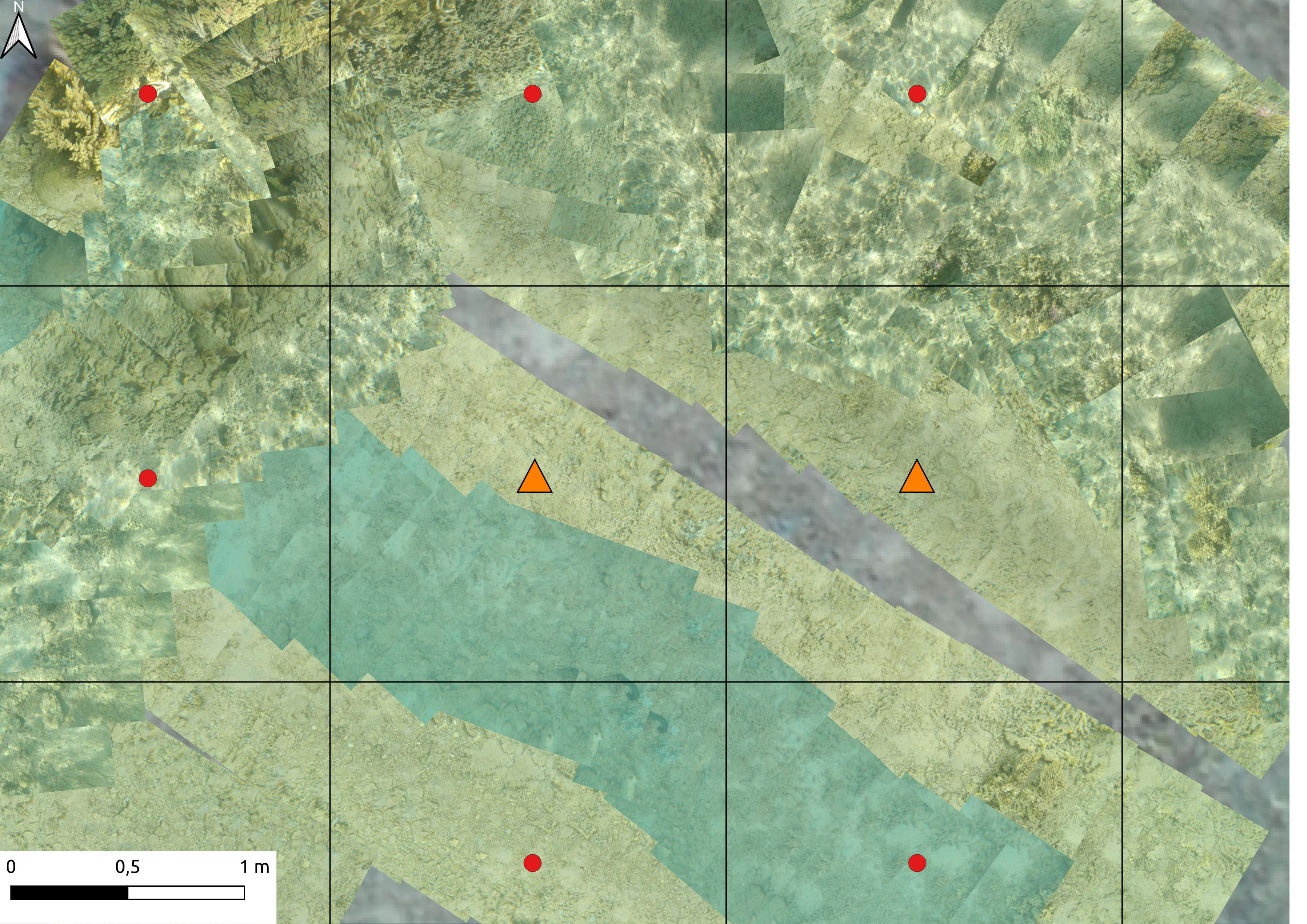}
        \caption{}
        \label{fig:uncovered_tile}
    \end{subfigure}
    \caption{Examples of useless tiles extracted from the aerial orthophoto of the \textit{Saint Leu} lagoon in Reunion Island: (a) Example of a tile extracted from the aerial orthophoto of the \textit{Saint Leu} lagoon in Reunion Island, with a high percentage of black pixels (b) Example of a group of tiles extracted from the aerial orthophoto of the \textit{Saint Leu} lagoon in Reunion Island, with corresponding underwater images. 
    The tiles in the middle do not have enough coverage of underwater images}
    \label{fig:useless_tiles}
\end{figure}

The second step is to filter out useless tiles. 
This is done by removing tiles with a high percentage of black pixels (due to errors in SfM processing) and tiles with no corresponding underwater images. 
Examples of such tiles are shown in Figure \ref{fig:useless_tiles}. 
In Figure \ref{fig:black_tile} we can see an example of a tile extracted from the aerial orthophoto of the \textit{Saint Leu} lagoon in Reunion Island with a high percentage of black pixels.

\subsubsection{Footprint calculation and tile coverage assessment} 
\label{subsubsec_footprint}

\begin{figure*}[ht]
    \centering
    \includegraphics[width=0.5\textwidth]{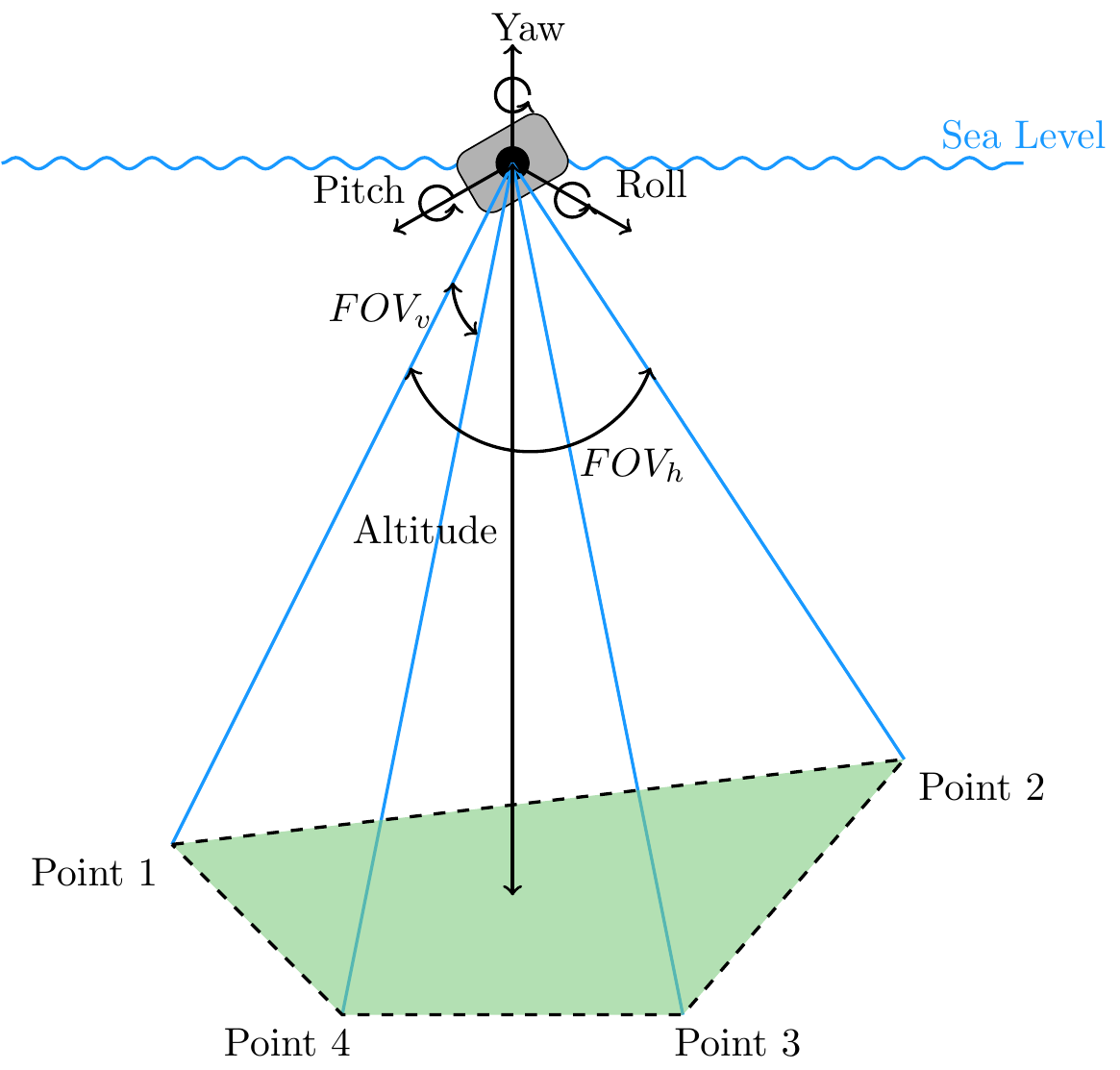}
    \caption{Footprint calculation of underwater images based on echosounder data, camera field of view and ASV angles.}
    \label{fig:footprint}
\end{figure*}

The third step involves associating underwater predictions with aerial tiles.
This assignment is achieved by identifying underwater images whose camera position centre falls within the boundaries of the aerial tiles.
After associating underwater images to aerial tiles, the next step is to compute the footprint of each underwater image to filter out aerial tiles with not enough underwater coverage.
In Figure \ref{fig:footprint}, we outline the process to calculate the footprint of underwater images based on data from ASV sensors.
Using bathymetric data from the echosounder, we measure the distance between the camera and the seabed, which determines the scale of the area captured in each image.
The camera orientation in the XYZ axis plane is defined by the roll, pitch, and yaw angles, which determine how the field of view (FOV) is directed relative to the seafloor.
Finally the FOV, divided into horizontal ($FOV_h$) and vertical ($FOV_v$) angles, defines the area visible to the camera. 
By projecting these angles down to the seafloor, we calculate the intersection points, forming a polygonal footprint that represents the region covered by the image. 
This footprint is necessary to associate each underwater image with a specific area of the seafloor.
Merging the footprints of all underwater images associated with a tile, we obtain the union of the footprints, which represents the area covered by the underwater images associated with the tile.

This allows us to filter out tiles with not enough underwater coverage. An example of such a tile is shown in Figure \ref{fig:uncovered_tile}, where a group of tiles extracted on the same orthophoto is shown. 
Tiles whose center is represented by a red point are classified as useful, since they are completely covered by underwater images.
On the contrary, tiles represented by an orange triangle are classified as useless, since they do not have enough coverage of underwater images.

\subsubsection{Transforming underwater predictions into aerial annotations} 
\label{subsubsec_transforming_predictions}

The fifth step is to transform underwater predictions into aerial annotations.
The trivial approach would be to associate the presence of a class $c$ in a tile $t$ if at least one underwater image associated with the tile is predicted as belonging to the class $c$  by the teacher model. 
Or, in other words, if class $c$ is not predicted as being absent on all underwater images associated with tile $t$, which can be formulated as:

\begin{equation}
    \forall c \in \mathcal{C}, \, \forall t \in \mathcal{T}, \quad I(y_c=1\;|\;t) = 1 - \prod_{x \in \mathcal{X}(t)} \left[1- h^{\text{teacher}}_c(x)\right]
\label{eq:presence_trivial}
\end{equation}

where :
\begin{itemize}
    \item $\mathcal{C}$ is the set of classes described in Section \ref{susubsec_aerial_dataset}
    \item $\mathcal{T}$ is the set of tiles
    \item $y_c \in \left\{0;1\right\}$ is the binary label associated with the presence/absence of class $c$
    \item $I(y_c=1\;|\;t) \in \left\{0;1\right\}$ is a binary 
    function indicating the presence or absence of class $c$ in tile $t$. 
    \item $\mathcal{X}(t)$ is the set of underwater images associated with tile $t$
    \item $h^{\text{teacher}}_c(x) \in \left\{0;1\right\}$ is the binary prediction associated with the presence/absence of class $c$ in underwater image $x$
\end{itemize}

The drawback of this approach is that it does not consider the footprint of underwater images, tending to overestimate the presence probability of a class in a tile.

A more realistic approach, since not all underwater images footprint fall entirely within the boundaries of a specific tile, needs to compute the intersection between the underwater image footprint and the corresponding tile.
This allows us to give more weight to underwater images that are completely within a tile and less weight to underwater images that are only partially within a tile.

\begin{figure*}[ht]
    \centering
    \begin{subfigure}[t]{0.45\textwidth}
        \centering
        \includegraphics[width=\textwidth]{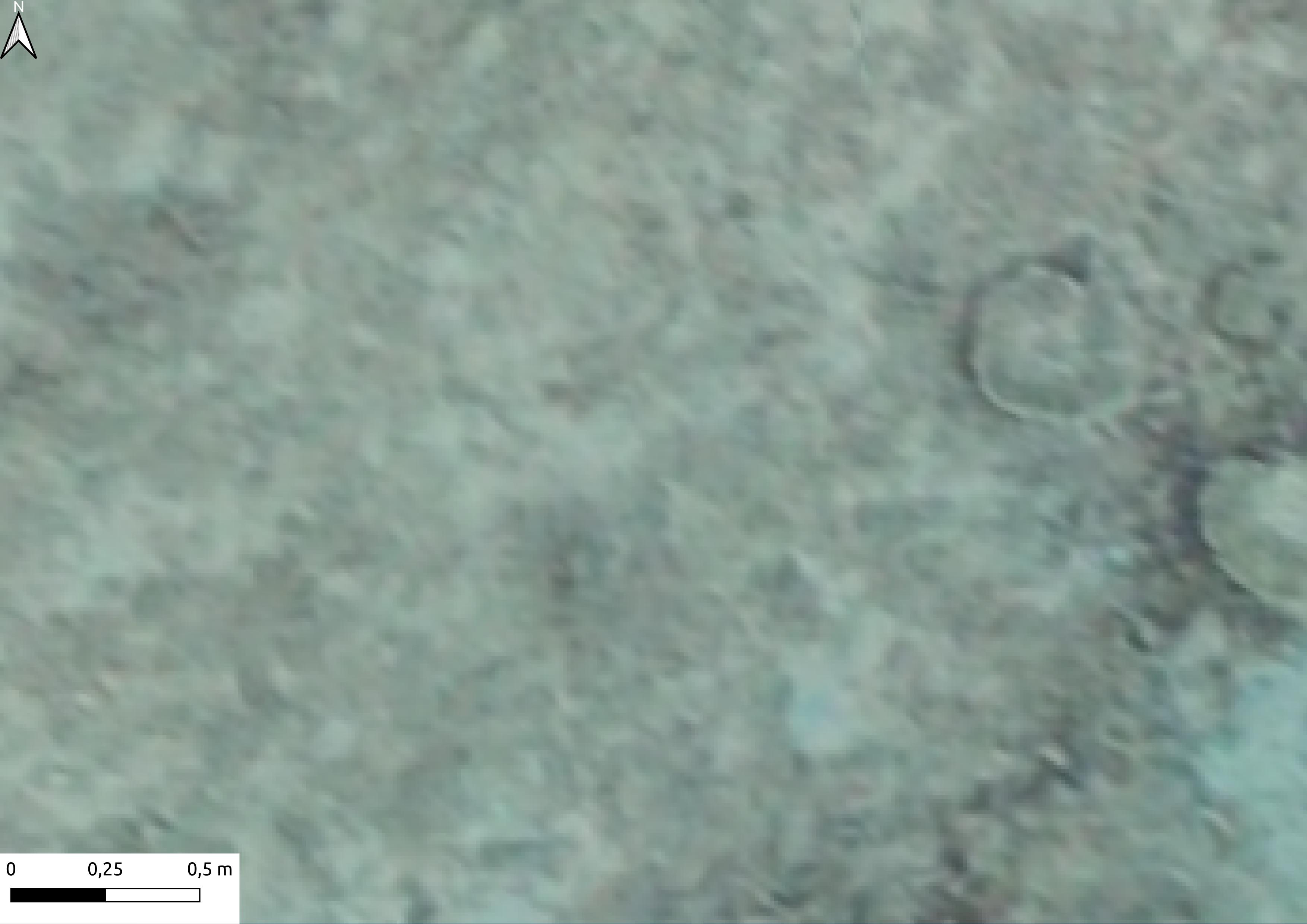}
        \caption{}
        \label{fig:ortho_tabulaire}
    \end{subfigure}
    \hfill
    \begin{subfigure}[t]{0.45\textwidth}
        \centering
        \includegraphics[width=\textwidth]{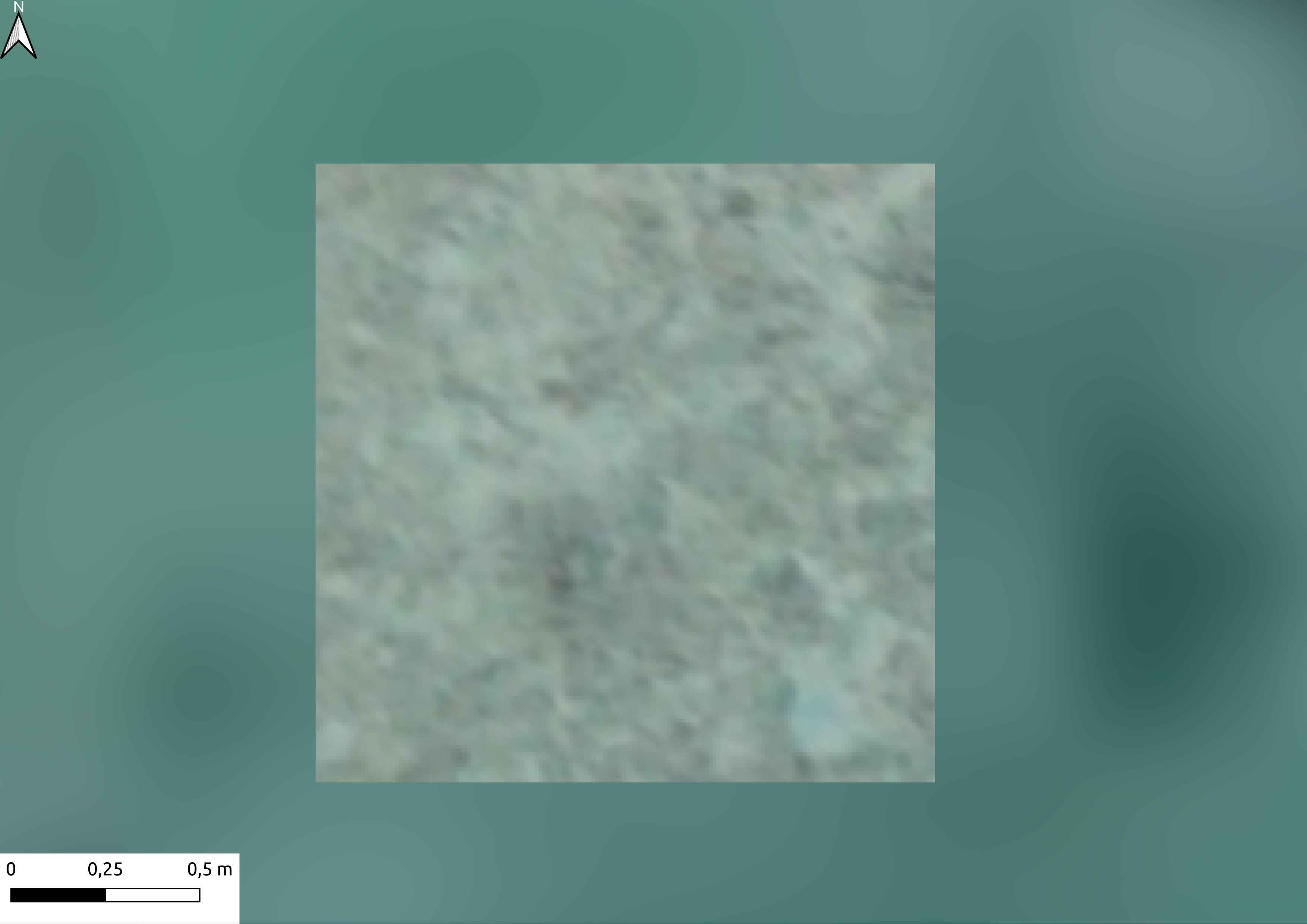}
        \caption{}
        \label{fig:tile_tabulaire}
    \end{subfigure}

    \vskip\baselineskip 
    
    \centering

    \begin{subfigure}[t]{0.45\textwidth}
        \centering
        \includegraphics[width=\textwidth]{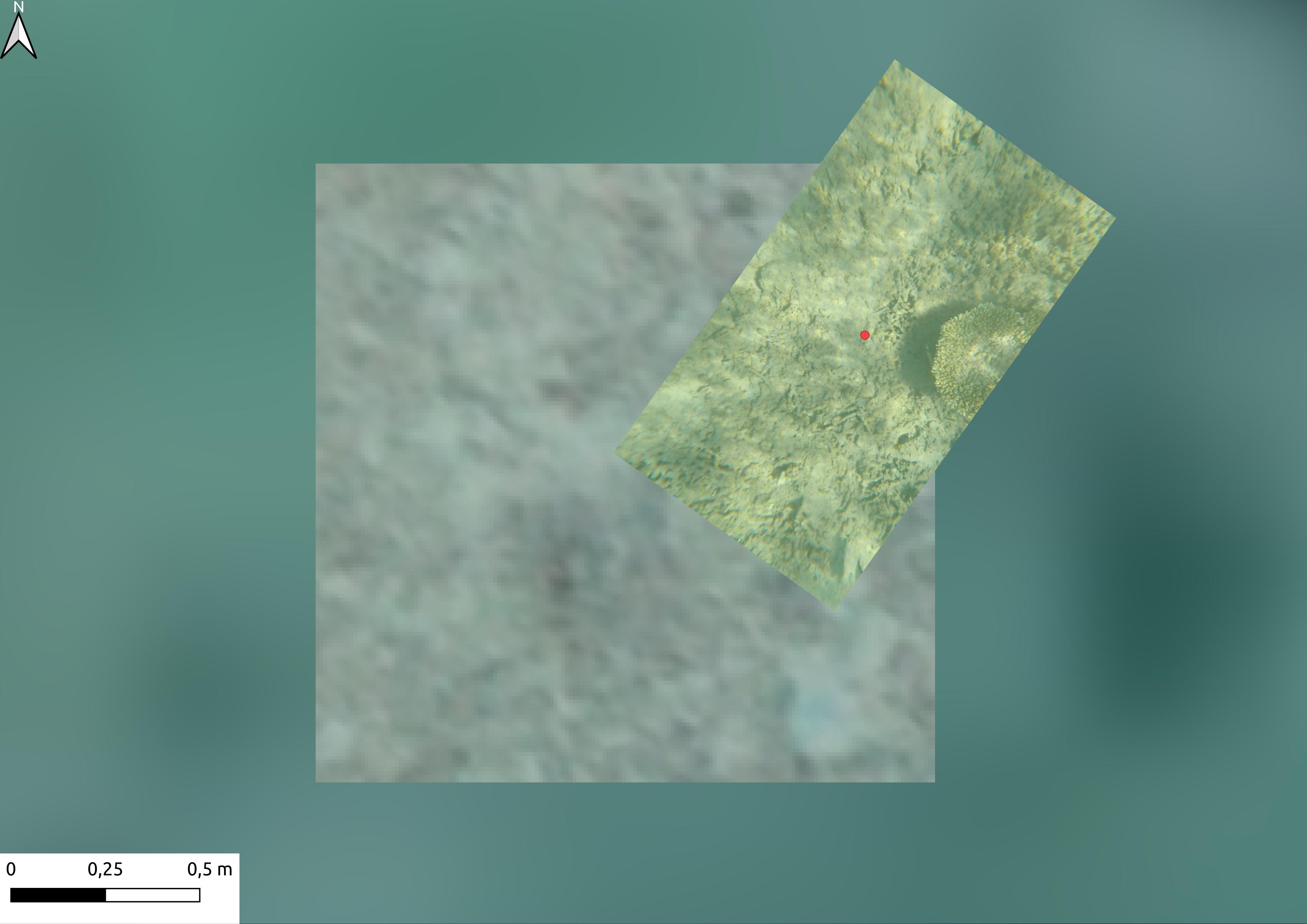}
        \caption{}
        \label{fig:tile_tabulaire_thumbnail}
    \end{subfigure}
    \hfill
    \begin{subfigure}[t]{0.45\textwidth}
        \centering
        \includegraphics[width=\textwidth]{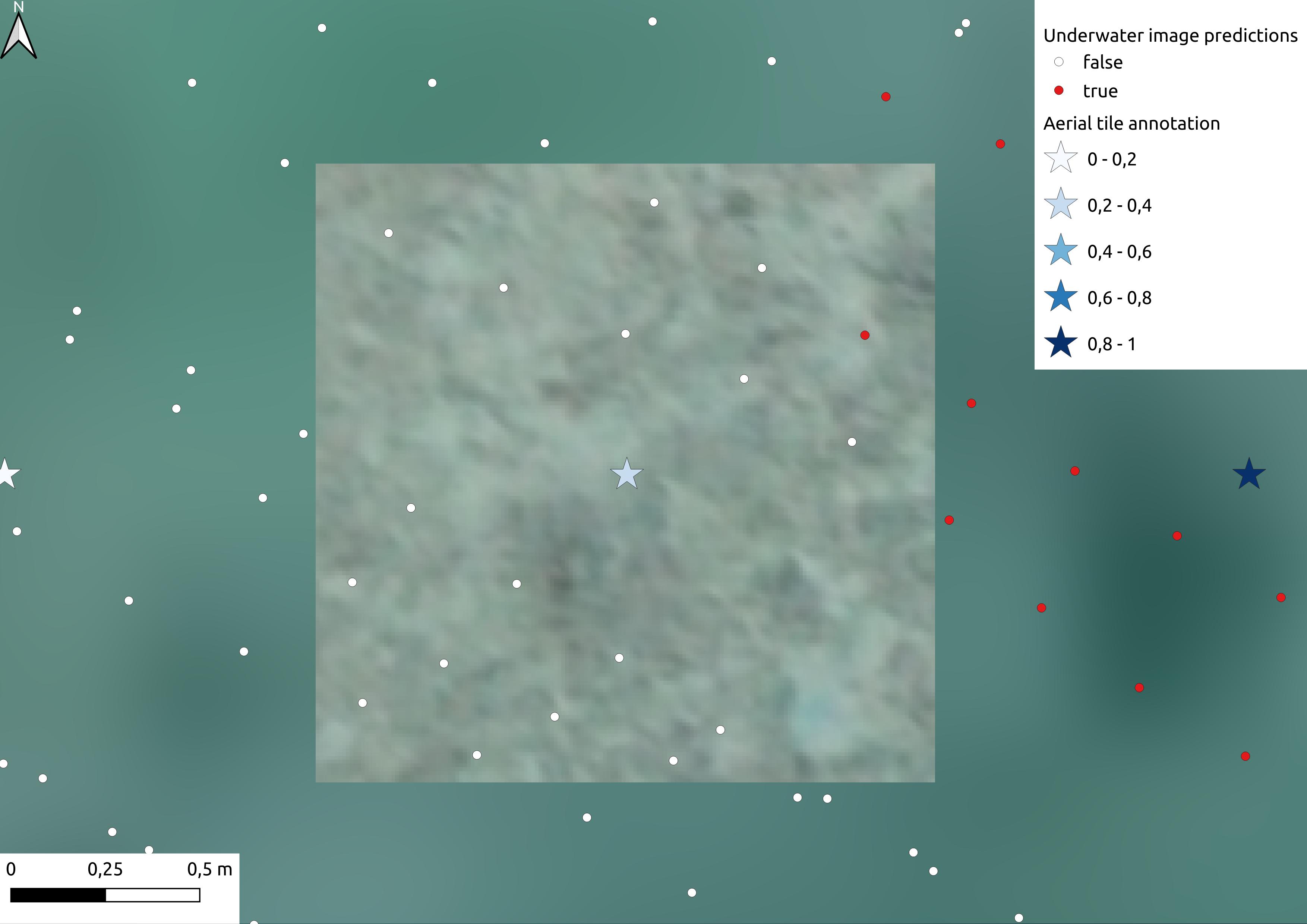}
        \caption{}
        \label{fig:tile_tabulaire_predictions}
    \end{subfigure}
    \caption{Example of the upscaling process from underwater predictions to aerial annotations, depending on the intersection between underwater images footprint and drone tiles, the probability of presence of a class in a tile can be mitigated: (a) Aerial orthophoto of the \textit{Trou d'eau} lagoon in Reunion Island (in the right part of the image a colony of $\AcroporeT$ corals is visible), (b) Drone tile extracted from the aerial orthophoto in Figure \ref{fig:ortho_tabulaire} (the $\AcroporeT$ corals are not visible in the tile), (c) Superposition of the underwater image in the top right corner of the tile and the drone tile (the $\AcroporeT$ coral colony is outside the drone tile), (d) Underwater predictions associated with the tile in Figure \ref{fig:tile_tabulaire} (the $\AcroporeT$ corals is predicted on the top right underwater image of the tile, but weakly predicted on the tile)}
    \label{fig:upscaling_annotations}

\end{figure*}

The orthophoto in Figure \ref{fig:upscaling_annotations} gives an example with the corresponding predictions on underwater images.
In the right part of Figure \ref{fig:ortho_tabulaire}, a colony of $\AcroporeT$ corals is visible.
Proceeding with tile extraction from the orthophoto, we obtain the tile in Figure \ref{fig:tile_tabulaire}.
Since the $\AcroporeT$ corals do not fall within the tile, we would like that, after computing the tile annotation starting from underwater predictions, the probability for the class $\AcroporeT$ associated with this tile will be weak.
Unfortunately, it may happen that underwater images that have the center within the tile (but not all the footprint) include classes that are outside the tile bounds: as shown in Figure \ref{fig:tile_tabulaire_thumbnail}, where a part of the $\AcroporeT$ coral colony is visible in the right part of the underwater image.
In these cases, weighting underwater predictions based on the intersection between the underwater image footprint and the tile allows reducing the impact of these images on the final aerial annotations.
This is shown in Figure \ref{fig:tile_tabulaire_predictions} where predictions on underwater images are represented with circles on a red ramp and aerial annotations are represented with stars on a blue ramp.
Even if in the underwater image in Figure \ref{fig:tile_tabulaire_thumbnail} on the right of the tile the presence of the $\AcroporeT$ class is predicted, the overlap between the underwater image and the aerial tile is weak.
Consequently, the probability of presence of the $\AcroporeT$ class on the tile is mitigated: ending up with an annotation of 0.4 (while the blue star on the tile just on the right indicates a probability of presence of 0.98).

To take account of the intersection between the underwater image footprint and the tile bounds, we can consider that the probability of presence is proportional to the intersection area relative to the underwater image area.   
Therefore, we can modify Equation \ref{eq:presence_trivial} as follows:

\begin{equation}
    \forall c \in \mathcal{C}, \, \forall t \in \mathcal{T}, \quad P(y_c=1\;|\;t) = 1 - \prod_{x \in \mathcal{X}(t)} \left[ 1 - \frac{s(x \cap t)}{s(x)}\;h^{\text{teacher}}_c(x)\right]
\label{eq:presence}
\end{equation}

where:
\begin{itemize}
    \item $P(y_c=1\;|\;t) \in [0,1]$ is the probability of presence of class $c$ in tile $t$
    \item \(s(x)\) is the area of the underwater image $x$
    \item \(s(x \cap t)\) is the area of intersection between the underwater image $x$ and the tile $t$
\end{itemize}

The product over all underwater images gives the probability that class $c$ is absent in all underwater images associated with the tile $t$.\\
\\
To get a better estimation of the presence probability, we can also take into account the confidence of the teacher model. Therefore, we can replace the binary output of the classifier $h^{\text{teacher}}_c(x)$ by the probabilistic output $p_{\text{teacher}}(y_c =1 \; | \; x)$ leading to:
\begin{equation}
    \forall c \in \mathcal{C}, \, \forall t \in \mathcal{T}, \quad P(y_c=1\;|\;t) = 1 - \prod_{x \in \mathcal{X}(t)} \left[ 1 - \frac{s(x \cap t)}{s(x)}\; p_{\text{teacher}}(y_c =1 \; | \; x)\right]
\label{eq:presence_adapted}
\end{equation}
where:
\begin{itemize}
    \item $p_{\text{teacher}}(y_c =1 \; | \; x) \in [0,1]$ is the probabilistic output of the teacher model for class $c$ in the underwater image $x$, obtained through a sigmoid function on top of the final layer of the model.
    If the model is trained with the binary cross-entropy loss function (as in our experiments), the output is asymptotically converging to the true conditional probability that class $c$ is present in image $x$  \cite{lorieul2020uncertainty}.
\end{itemize}
It is worth noting that when the probability $p_{\text{teacher}}(y_c =1 \; | \; x)$ is equal to 1, then Equation \ref{eq:presence_adapted} is equivalent to Equation \ref{eq:presence}. 
But in the general case, it is comprised in the interval $]0,1[$. In the literature related to knowledge distillation \cite{gou2021knowledge}, such probabilistic labels passed to the student model are often called \textit{soft labels} in opposition to hard labels such as the one in Equation \ref{eq:presence_trivial}. 
Training the student model on soft labels rather than hard labels enables a better transfer of information from the teacher to the student. 
The probability of presence actually captures valuable information on the uncertainty of the teacher model, and allows us to place less weight on ambiguous cases in the loss function of the student.

\subsubsection{Aerial dataset} 
\label{susubsec_aerial_dataset}
Following the upscaling process detailed in Section \ref{subsubsec_upscaling_predictions}, starting from two aerial orthophotos of the \textit{Trou d'eau} and \textit{Saint-Leu} lagoons in Reunion Island measuring 189,682 m\(^2\) and 204,748 m\(^2\) respectively, we ended up with 4,911 and 6,832 
annotated tiles respectively for a total of 11,743 annotated tiles.

Since the upscaling process implies a loss in the image resolution, we made some changes about the classes to be predicted:
\begin{enumerate}
    \item The first change was to merge \textit{Algae} classes into a single class called \textit{Algae}, indeed distinguishing between the different types of algae ($\AlgaeA$, $\AlgaeD$, $\AlgaeL$ and $\AlgaeS$) is a task that requires a higher resolution than the one we have\footnote{In the case of Equation \ref{eq:presence} this was done by assigning $h^{\text{teacher}}_{\textit{Algae}}(x)=1$ if at least one type of algae ($\AlgaeA$, $\AlgaeD$, $\AlgaeL$ and $\AlgaeS$) was predicted as present on the fine scale image $x$, otherwise $h^{\text{teacher}}_{\textit{Algae}}(x)=0$.
    In the case of Equation \ref{eq:presence_adapted} this was done by assigning to $p_{\text{teacher}}(y_{\textit{Algae}} =1 \; | \; x)$ the maximum between all the probabilities predicted by the underwater model for algae classes ($\AlgaeA$, $\AlgaeD$, $\AlgaeL$ and $\AlgaeS$).}.

    \item The second change was to remove underwater classes that do not have a corresponding aerial class: $\Blurred$ images (an underwater blurred image does not imply a blurred aerial image) and $\Homo$ (since human body parts in underwater images do not imply human body parts in aerial images).
    \item The third change was to remove underwater classes that are not relevant for the aerial images, i.e. $\Fish$, $\SeaC$ and $\SeaU$. 
    The first two classes, even if visible in some aerial images, were removed because underwater and aerial images are not taken at the same time, so that the presence of a sea cucumber or a fish in an underwater image does not imply the presence of those organisms in the corresponding aerial image.
    The last one was removed since those organisms are not visible at all in aerial images.
\end{enumerate}

Finally, we retained only classes for which there was a sufficient number of annotations.
Thus, removing classes that have less than 200 annotations in the aerial dataset, we ended up with 12 classes:

\begin{itemize}

    \item Coral
        \begin{multicols}{3}
        \begin{enumerate}
            \item $\AcroporeB$
            \item $\AcroporeD$
            \item $\AcroporeT$
            \item $\Dead$
            \item $\NoAcroporeE$
            \item $\NoAcroporeM$
            \item $\Millepore$
            \item $\NoAcroporeSu$
        \end{enumerate}
        \end{multicols}

    \item Habitat
        \begin{multicols}{3}
        \begin{enumerate}
            \item $\Rock$
            \item $\Rubble$ 
            \item $\Sand$ 
        \end{enumerate}  
        \end{multicols}

    \item Other Organisms
        \begin{multicols}{3}
        \begin{enumerate}
            \item \textit{Algae} 
        \end{enumerate} 
        \end{multicols}
         
\end{itemize}

\subsubsection{Aerial deep learning model (student model)} 
\label{subsubsec_aerial_deep_learning_model}

To train the student model with soft labels, we use the Binary Cross-Entropy (BCE) with logits loss function. 
This loss measures the divergence between the predicted logits of the student model and the soft labels $P(y_c=1\;|\;t) \in [0,1]$ generated by the teacher model. 
Specifically, the loss for class $c$ in tile $t$ is given by:

\begin{equation}
    \mathcal{L}_{\text{BCE}}(t, c) = - \Big[ P(y_c = 1 \;|\; t) \cdot \log(p_{\text{student}}(y_c = 1 \;|\; t)) + (1 - P(y_c = 1 \;|\; t)) \cdot \log(1 - p_{\text{student}}(y_c = 1 \;|\; t)) \Big] 
\label{eq:bce_loss}
\end{equation}

where:
\begin{itemize}
    \item $P(y_c=1\;|\;t) \in [0,1]$ is the soft label provided by the teacher model for class $c$ in tile $t$, as described in equations \ref{eq:presence} and \ref{eq:presence_adapted}
    \item $p_{\text{student}}(y_c =1 \; | \; t) \in [0,1]$ is the probabilistic output of the student model for class $c$ in tile $t$, obtained through a sigmoid function on top of the final layer of the model.
\end{itemize}

To maintain consistency with underwater predictions, we used the same architecture for the student model as the one used for the teacher model (i.e. the DinoV2 model \cite{oquab_dinov2_2023}).
\\
The only difference is that, since the underwater model was trained with binary values and the aerial model has to be trained on probabilities, when computing evaluation metrics during the training process we cannot use the accuracy, precision, recall and F1-score metrics. 
Instead, we will compute the Root Mean Squared Error (RMSE), the Mean Absolute Error (MAE) and the Kullback-Leibler (KL) divergence metrics.

\subsection{Test zone and model evaluation}
\label{subsec_test_zone}

\begin{figure*}[ht]
    \centering
    \includegraphics[width=0.5\textwidth]{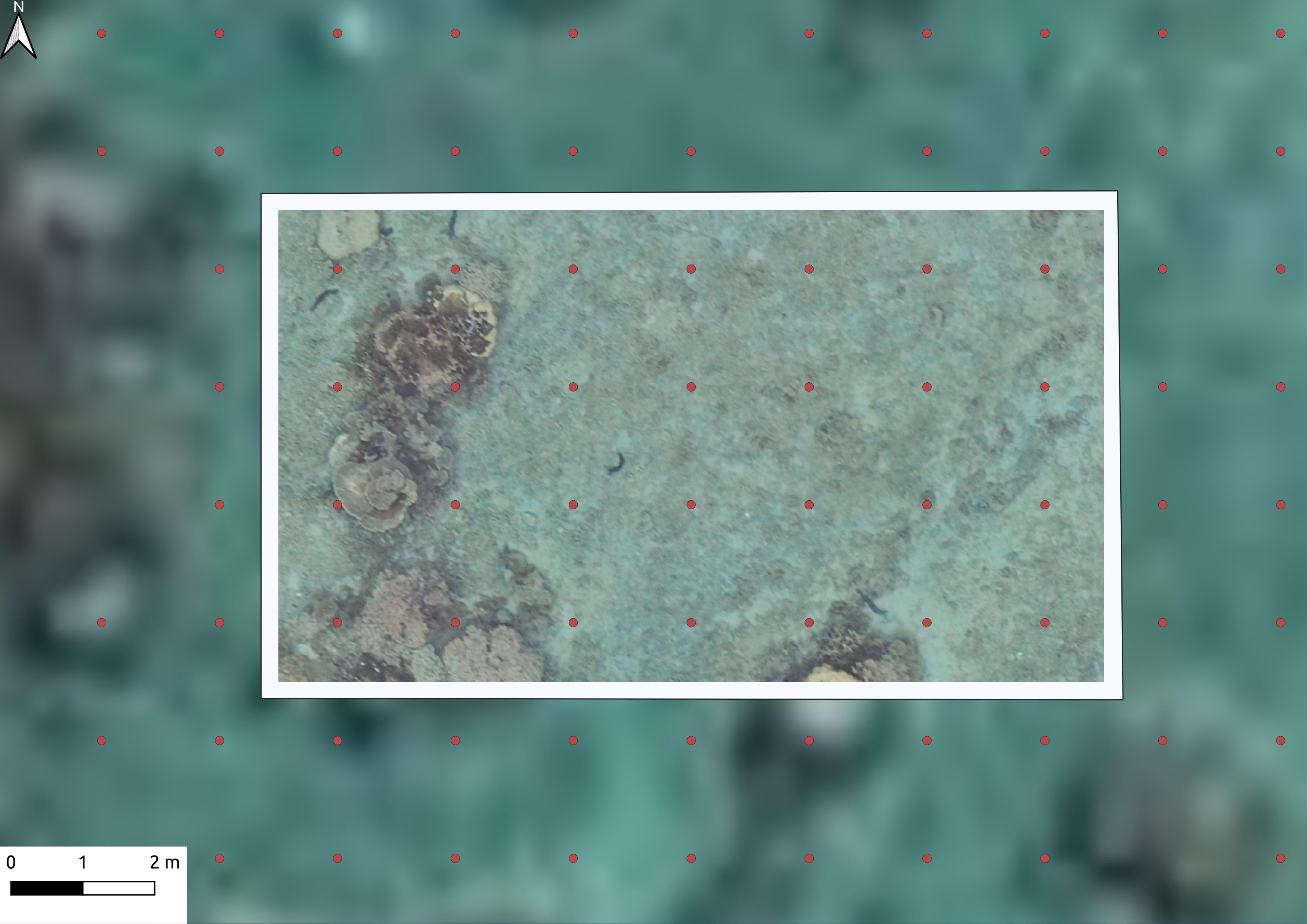}
    \caption{Test zone within the \textit{Trou d'eau} lagoon, selected for model evaluation. The area measures $63 m^2$ and comprises 194 underwater images corresponding to 28 aerial tiles.}
    \label{fig:test_zone}
\end{figure*}

To evaluate the performance of the aerial deep learning model, we selected a test zone within the \textit{Trou d'eau} lagoon, see Figure \ref{fig:test_zone}. 
This area was chosen due to its diverse composition of coral morphotypes, habitats, and other marine organisms, representing a challenging environment for model validation.
The test zone comprises 194 underwater images, corresponding to 28 aerial tiles, for a total of $63 m^2$.

The annotation process for the aerial tiles is carried out as follows:

\begin{enumerate} 
    \item For each aerial tile, underwater images with centroids located within the tile boundaries are identified
    \item These underwater images are then projected in QGIS to assess the portions of each image that intersect the aerial tile boundaries 
    \item Each cropped underwater image is manually annotated with fine-grained precision
    \item As a result, each aerial tile is annotated with a level of detail comparable to that of underwater imagery and is therefore considered as ground truth data
\end{enumerate}

These aerial tiles were not used during model training, ensuring an unbiased evaluation of the aerial model.


Both equations \ref{eq:presence} and \ref{eq:presence_adapted} were used to generate aerial annotations starting from underwater predictions in the test zone.
The generated annotations were then compared with ground truth data to evaluate the goodness of the upscaling process using the AUC (Area Under the Curve) metric, which is commonly used metric in the evaluation of Species Distribution Models (SDMs) \cite{5e86dc5ce7094d39ba95dd74ff975885}.

Finally, in order to evaluate the aerial model, we compared the predictions on the test zone with the ground truth data using the AUC metric.

\section{Results} 
\label{sec_results}

\subsection{Upscaling process evaluation}
\label{subsec_upscaling_process_evaluation}

In order to evaluate the upscaling process, we compared the generated annotations with ground truth annotations in the geospatial test zone in the \textit{Trou d'eau} lagoon.
We first evaluated the quality of the soft labels generated with our methods described in Equations \ref{eq:presence} and \ref{eq:presence_adapted}. 
As they provide a presence probability for each class, we can actually measure their AUC on the ground truth annotations. 
With a value of 0.9211 for annotations generated through Equation \ref{eq:presence} and 0.9251 for annotations generated through Equation \ref{eq:presence_adapted}, both methods show a high level of accuracy in transferring information across scales.

To further evaluate both methods, we then measured the  performance of the student model trained with either method.
In the following, we will call \texttt{Model\_spatial\_only} the model trained from annotations generated through Equation \ref{eq:presence} and \texttt{Model\_distilled} the model trained from the annotations generated through Equation \ref{eq:presence_adapted}. Only the second model integrates information about the teacher's model confidence (= knowledge distillation). The first model integrates the hard labels predicted by the teacher and the spatial coverage. 
We first looked at the evaluation metrics measured on the soft labels themselves (using the random test set). 
The results are shown in Table \ref{tab:model_comparison}.

\begin{table}[ht]
    \centering
    \resizebox{0.5\columnwidth}{!}{
    \begin{tabular}{lccc}
    \toprule
    Model & RMSE & MAE & KL Divergence \\
    \midrule
    \texttt{Model\_spatial\_only} & 0.2019 & 0.1446 & 0.9802 \\
    \texttt{Model\_distilled} & \textbf{0.1546} & \textbf{0.1143} & \textbf{0.3931} \\
    \bottomrule
    \end{tabular}
    }
    \caption{Comparison of \texttt{Model\_spatial\_only} and \texttt{Model\_distilled} on various performance metrics}
    \label{tab:model_comparison}
\end{table}

The results show that \texttt{Model\_distilled} trained using knowledge distillation (i.e. with Equation \ref{eq:presence_adapted}) allows a better prediction of the soft labels than \texttt{Model\_spatial\_only} trained without knowledge distillation (i.e. Equation \ref{eq:presence}) on all metrics. 
This means that the information they contain is more predictable from the aerial image contents. 

Finally, comparing the predictions generated with both \texttt{Model\_spatial\_only} and \texttt{Model\_distilled} on the ground truth data of the geospatial test zone in the \textit{Trou d'eau} lagoon, we obtain an AUC of  0.7753 and 0.7952 respectively.
This confirms that the best upscaling method is the one using knowledge distillation (Equation \ref{eq:presence_adapted}) and that high AUC values can be achieved by the aerial model using this method.

\subsection{Prediction maps}
\label{subsec_prediction_maps}
Once the aerial (student) model is trained, we can use it in inference mode to generate high resolution maps of large areas. 
In particular, we ran it on 20,027 tiles in the \textit{Trou d'eau} lagoon and 61,059 tiles in the \textit{Saint-Leu} lagoons.
For each tile, we used the output of the student model as the probability of presence of each class and then we generated prediction maps for each class.

\begin{figure}[ht]
    \centering
    \begin{subfigure}[t]{0.45\textwidth}
        \includegraphics[width=\textwidth]{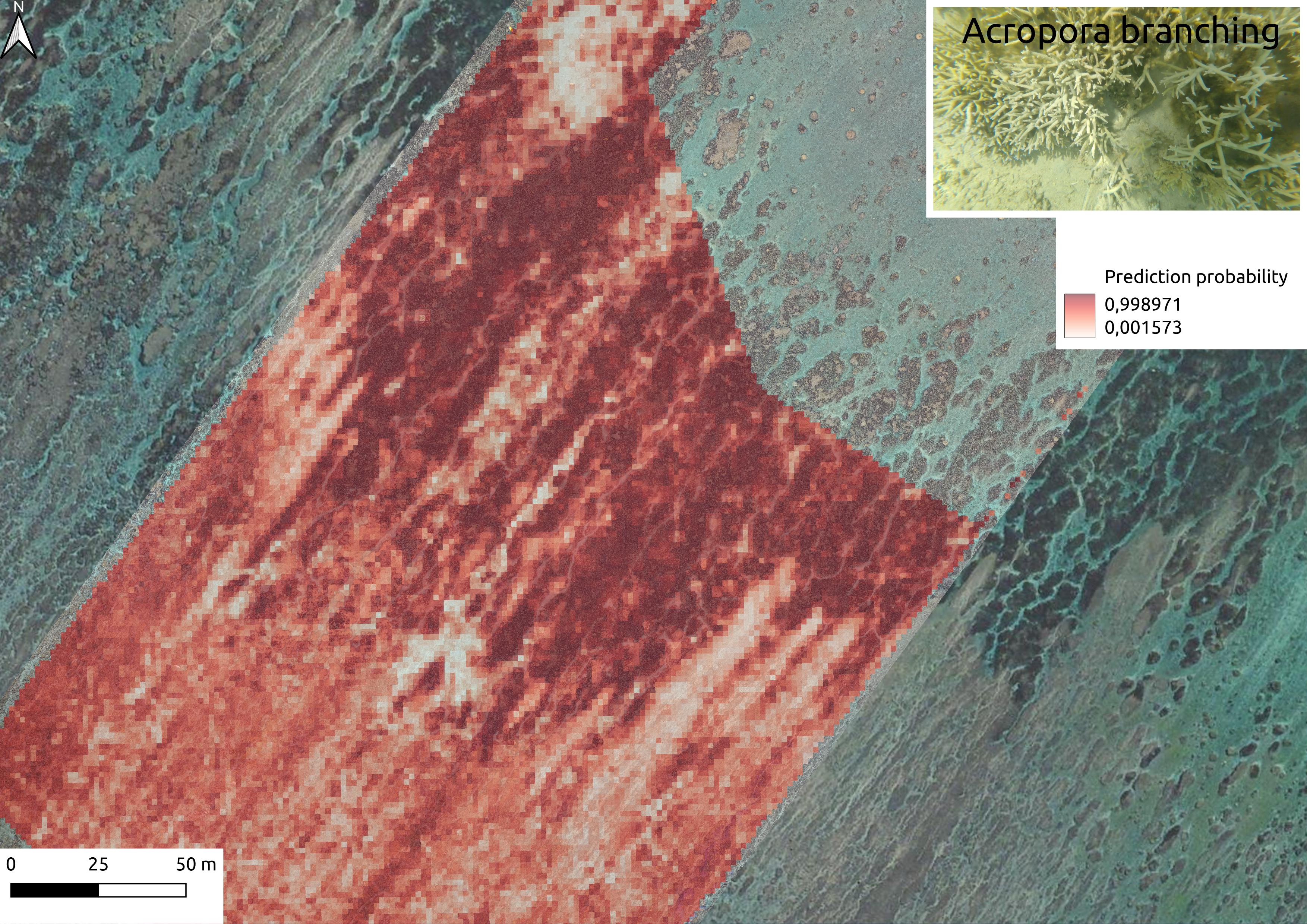}
        \caption{}
        \label{fig:predictions-acr-br}
    \end{subfigure}
    \hfill
    \begin{subfigure}[t]{0.45\textwidth}
        \includegraphics[width=\textwidth]{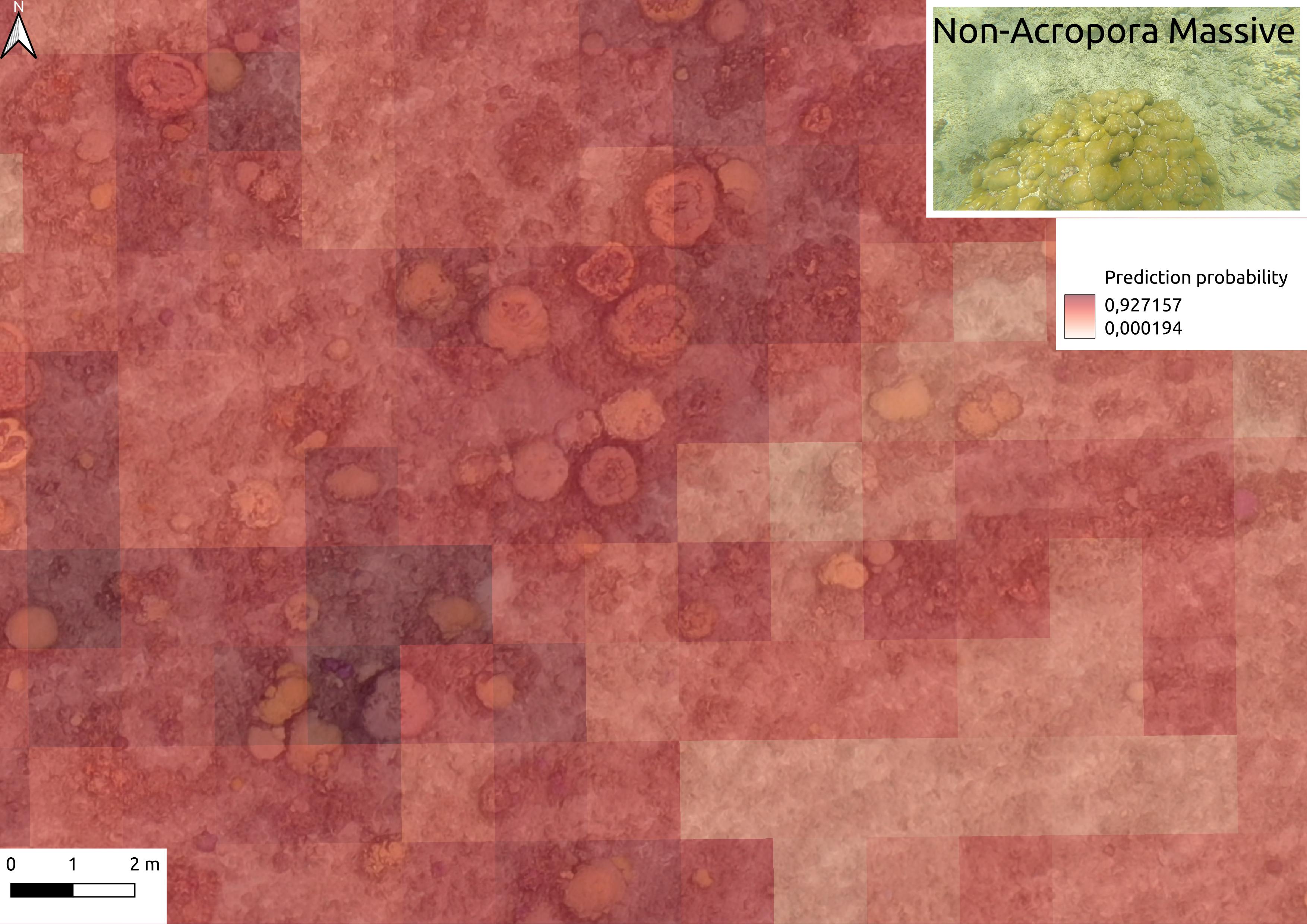}
        \caption{}
        \label{fig:predictions-porites}
    \end{subfigure}
    \caption{Prediction maps generated by the aerial model superposed on the aerial orthophoto: (a) Prediction map for the $\AcroporeB$ class in the \textit{Trou d'eau} lagoon. Red patches indicate the presence of corals.
    (b) Zoomed-in prediction map for the $\NoAcroporeM$ class in the \textit{Saint-Leu} lagoon.
    Blue patches indicate the presence of corals.}
    \label{fig:predictions}
\end{figure}

In Figure \ref{fig:predictions} we show two examples of prediction maps generated by the aerial model for the 
\\
$\AcroporeB$ class in the \textit{Trou d'eau} lagoon and the $\NoAcroporeM$ class in the \textit{Saint-Leu} lagoon.
The granularity of the prediction raster is fixed at 1.5 m x 1.5 m, which is the same as the aerial tiles.
Different spatial scales are shown in Figure \ref{fig:predictions-acr-br} and Figure \ref{fig:predictions-porites}, in order to highlight the model's ability to cover large areas while still being able to capture fine-scale details.

\section{Discussion} 
\label{sec_discussion}
This study demonstrates the potential of combining underwater and aerial imagery to improve monitoring and management of coral reef ecosystems. 
Although previous research has highlighted the advantages of using both imaging techniques, this work, to our knowledge, is the first to synergize AI models across different scales. 
By using high-resolution underwater AI predictions to train a larger-scale aerial model, we ensure the precision of underwater analysis while extending it to cover a broader reef area. 
This multi-scale approach has the potential to advance marine monitoring, and it can also be applied to other fields such as agriculture, forestry, urban planning, and so forth.

\subsection{Transferring information across scales: model independence and flexibility} 
\label{subsection:transferring_information_across_scales}

\subsubsection{Upscaling process evaluation}
\label{subsubsection_upscaling_process_evaluation}

While having a well-performing fine-scale model is needed for reliable medium-scale annotations,
working on the model architecture in order to gain a few percentage points of accuracy is not the core of our methodology.
The primary objective of our workflow is to train a medium-scale student model to mimic the behaviour of a fine-scale teacher model, without the need to reannotate medium-scale images. 

Two different techniques were used to generate aerial annotations starting from underwater predictions: the first one was based on presence/absence values obtained by thresholding underwater predictions (Equation \ref{eq:presence}), while the second one was based on knownledge distillation, i.e., on integrating the probability values predicted by the underwater model in the soft labels passed to the student (Equation \ref{eq:presence_adapted}).

To compare and validate those methodologies, we evaluated them based on underwater predictions with ground truth data in the test zone in the \textit{Trou d'eau} lagoon. 
The distillation-based method appeared to be the best with a high value of the AUC metric (0.9251).  
This indicates that it is a reliable method for transferring information from fine-scale to medium-scale images, allowing for a more nuanced estimate of class presence compared to using binary predictions \ref{eq:presence}.

Indeed, the soft labels generated by the teacher model provide a more accurate representation of class presence compared to the hard targets, since they capture the probability distribution across all possible classes, offering continuous values between 0 and 1. This approach allows the student model to better learn from the teacher model's uncertainty, leading to improved generalization capabilities and performances \cite{journals/corr/HintonVD15}.
This result is consistent with the performance of the aerial model trained with annotations generated by Equation \ref{eq:presence_adapted}, which outperformed the model trained with annotations generated by Equation \ref{eq:presence} on all metrics.

A key advantage of this methodology is its model independence. 
If a more accurate or advanced model becomes available in the future (thanks to new annotated images or improved algorithms), our framework allows for easy adaptation.
By applying the new model to our existing fine-scale data, we can regenerate high-resolution predictions, which can then be seamlessly transferred to the medium scale. 
This adaptability ensures that the model continues to benefit from the most accurate fine-scale insights.

\subsubsection{Aerial model evaluation}
\label{subsubsection:model_evaluation}

The model's performance on the neural network test set underscores its robustness. 
As shown in Table \ref{tab:model_comparison}, the RMSE, MAE, and KL Divergence metrics are all low, indicating that the model's predictions closely match annotations on the test set.
The relatively low MAE compared to the RMSE (0.1143 vs. 0.1546) suggests that predictions on average deviate from the true values by less than 12\%, indicating a high level of accuracy.
Occasional larger discrepancies make the RMSE slightly higher, but the overall results are still very promising.
The high AUC value (0.7952) computed on the test zone further confirms the model's strong performance, indicating that the model can reliably generate probability estimates that closely align with the true distributions of the labels.

With regard to possible improvements concerning the deep learning model, as previously mentioned, in this study we used \textit{DinoV2} as a backbone: one of the SOTA (State Of The Art) computer vision models that currently performs the best on benchmark datasets.
Applying transfer learning, we take advantage of  the model’s strong generalization capabilities while fine-tuning it to address our specific problem.
However, we recognize that with the rapid advancements in artificial intelligence, our model may already be on the path to obsolescence \cite{ma2024surveyvisionlanguageactionmodelsembodied}. 
Continuously updating the model predictions to train the medium-scale model would enable us to incorporate the latest breakthroughs in AI, ensuring increasingly refined annotation quality over time.

\subsection{Georeferencing challenges in multi-scale monitoring} 
\label{subsection:georeferencing_challenges}
A framework that enables the transfer of information from fine-scale to a broader-scale imagery relies essentially on achieving precise alignment between the two layers of data.
As shown in Figure \ref{fig:upscaling_annotations} the benthic substrate can vary significantly across small distances, so that in order to upscale underwater predictions to aerial annotations, data need to be accurately georeferenced.

Using differential GPS technology is the first step to achieve precise positioning, but does not guarantee a centimetric accuracy.
To improve the ASV positioning, we used PPK techniques thanks to the CentipedeRTK network \footnote{\url{ https://docs.centipede.fr/}}, ending up with a centimetric accuracy in the ASV positioning.
Unfortunately, the position of the ASV is not the same as the position of the image, since waves can change the attitude of the ASV by tilting the direction of the camera from the vertical axis (see Figure \ref{fig:footprint}).
To correct this, we used the camera angles on the three axes (Roll, Pitch and Yaw) and the echosounder data to compute the footprint of underwater images.
Ending up with the latitude and longitude of the four corners defining the footprint on the seabed of each underwater image.
A check on the quality of the georeferencing of underwater images is then necessary in order to validate the image positioning accuracy with an unbiased approach.
In our case we chose to compare our data with data produced by the French National Institute of Geographic and Forest Information (IGN), which is a reference in the field of georeferencing. 
Thanks to a visual comparison between the two datasets, we were able to validate the accuracy of the fine scale georeferencing process.

Although using a differential GPS on the platform acquiring images is a significant advantage, it is neither a necessary nor a sufficient condition to achieve centimetric accuracy. 
For instance, our methodology demonstrated that combining PPK techniques, GCPs, and validation against an external reference such as IGN data can deliver the required accuracy for coral reef monitoring.
However, since our current setup involves UAVs without embedded differential GPS, precise aerial image alignment still depends on the manual collection of GCPs, making the process time-consuming and labor-intensive. 
Transitioning to UAVs equipped with embedded differential GPS could significantly streamline the data collection process. 
By applying PPK techniques directly to aerial images, the need for GCP collection could be minimized to a few validation points, reducing mission planning time and increasing efficiency. 
This advancement would enable faster and more scalable reef monitoring over large areas while maintaining accurate positioning.

\subsection{Expanding spatial coverage and species identification} 
\label{subsection:expanding} 

\subsubsection{Satellite imagery upscaling} 
\label{subsubsection:satellite} 
This study highlights several areas for future improvement. 
Although the classification accuracy was high across most classes, certain coral types remain challenging to differentiate at the aerial scale due to image resolution constraints. 
Addressing this limitation may involve refining aerial image resolution by reducing the flight altitude (respecting the regulations in force in the country where the data is collected), employing more advanced image processing techniques, using higher-quality drones / cameras with better sensors \cite{9999327} or even using hyperspectral cameras to capture more detailed information about the reef \cite{ROSSITER2020106789}.

Finally, mimicking the idea presented in this study, we could extend the methodology to the satellite scale.
In \cite{alvarez2021uav} the authors discuss the complementary nature of UAV and satellite data, pointing out that integrating these technologies can improve spatial and temporal resolution in remote sensing applications. 
Successful integration would allow rapid monitoring of large areas, significantly reducing the need for field data collection, as satellite imagery is often freely available and collected at regular intervals. 
Furthermore, access to a historical archive of satellite imagery provides a unique opportunity to study ecosystem evolution over past decades, while supporting long-term monitoring efforts in the future. 
This extended temporal and spatial coverage could greatly improve our understanding of ecological change on a global scale.

\subsubsection{Expanding to slow-moving species identification through synchronized imaging} 
\label{subsubsection:concombres} 

The collection of both fine-scale and aerial images simultaneously has the potential to enable the identification of certain benthic species, which would otherwise be challenging to recognise. 
To illustrate, slow-moving organisms such as sea cucumbers are frequently visible in aerial images (e.g., Figure \ref{fig:test_zone}). 
Given that these species move at a slow speed, synchronised collection of both image types would provide the temporal and spatial alignment necessary for passing the information from the fine-scale model to the medium-scale model.
Although this data collection approach imposes additional constraints compared to the method used in this study, where fine-scale and medium-scale data can be collected days or even months apart, it offers the advantage of providing additional ecological insights that would otherwise be inaccessible
\cite{conand:hal-01906874}.

\section{Conclusion} 
\label{sec_conclusion}
In this study, we presented a novel methodology for transferring information across scales in coral reef monitoring.
By combining high-resolution underwater imagery with medium-scale aerial data, we were able to train a deep learning model to predict benthic substrate composition over a large reef area.
Our approach leverages the strengths of both imaging techniques, ensuring the precision of fine-scale analysis while extending it to cover a broader reef area.
This multi-scale framework has the potential to revolutionize marine monitoring, providing a more comprehensive and efficient way to assess coral reef health.

Our methodology is not only innovative but also highly adaptable.
By training a medium-scale model to mimic the behaviour of a fine-scale model, we ensure that the system can easily incorporate new advances in AI without the need for data re-annotation.
This flexibility allows us to continuously improve the model’s predictions, ensuring that it remains at the cutting edge of coral reef monitoring.
Using standardized annotation protocols and adhering to FAIR (Findable, Accessible, Interoperable, and Reusable) data principles can broaden the range of ecosystems monitored with our methodology. 

Looking ahead, we see great potential for our methodology to be extended to the satellite scale.
By integrating UAV and satellite data, we can enhance spatial and temporal resolution in remote sensing applications, providing a more comprehensive view of coral reef ecosystems.
This extended coverage could greatly improve our understanding of ecological change on a global scale, supporting long-term monitoring efforts and contributing to the conservation of these vital marine ecosystems.

Moreover, the proposed upscaling methodology shows promise for applications in a number of different fields beyond the monitoring of coral reefs. 
In terrestrial environments, it could be used to support forestry management by extending detailed ground-based observations to regional scales. 
In a completely different domain, like in urban planning, ground-level observations of pedestrians, traffic or vegetation could inform city-wide analyses using aerial or satellite imagery.
These examples demonstrate the versatility of this approach, which could be applied across a range of different disciplines and environments.

\section{CRediT authorship contribution statement} 
\label{sec_author_contributions}
\textbf{Matteo Contini}: Conceptualization, Data curation, Methodology, Formal analysis, Writing – original draft.
\textbf{Victor Illien}: Conceptualization, Data curation, Methodology, Software, Writing – review and editing.
\textbf{Julien Barde}: Conceptualization, Data curation, Funding acquisition, Methodology, Supervision, Writing – review and editing.
\textbf{Sylvain Poulain}: Data curation, Methodology, Software.
\textbf{Serge Bernard}: Conceptualization, Data curation, Funding acquisition, Methodology, Supervision, Writing – review and editing.
\textbf{Alexis Joly}: Conceptualization, Data curation, Funding acquisition, Methodology, Project administration, Writing – review and editing.
\textbf{Sylvain Bonhommeau}: Conceptualization, Data curation, Funding acquisition, Methodology, Project administration, Writing – review and editing.

\section{Acknowledgement} 
\label{sec_acknowledgement}
The authors acknowledge the Pôle de Calcul et de Données Marines (PCDM) for providing DATARMOR storage, support services and computational resources. 
We extend our heartfelt thanks to Laurence Maurel, Cam Ly Rintz, Leanne Carpentier, Magali Duval, Laura Babet, Belen De Ana, Anne-Elise Nieblas, Arthur Lazennec, Victor Russias, Mervyn Ravitchandirane, Mohan Julien, Pierre Gogendeau, Thomas Chevrier and Justine Talpaert Daudon, involved in the annotation and collection of data for this study. 
Their contributions were indispensable to our research efforts. 
We also thank Emilien Alvarez for his valuable advice on drone imaging. 

\section{Funding} 
\label{sec_funding}
This work was supported by several projects: Seatizen (Ifremer internal grant), Plancha (supported by the Contrat de convergence et de transformation 2019-2022, mesure 3.3.1.1 de la Préfecture de la Réunion, France), IOT project (funded by FEDER INTERREG V and Prefet de La Réunion: grant \#20181306-0018039 and the Contrat de Convergence et de Transformation de la Préfecture de La Réunion), Ocean and Climate Priority Research Programme, FISH-PREDICT project (funded by the IA-Biodiv ANR project: ANR-21-AAFI-0001-01), B1-4SEA funded by Explorations de Monaco  and G2OI FEDER INTERREG V (grant \#20201454-0018095).

\section{Conflict of Interest statement} 
\label{sec_conflict}
The authors declare no conflict of interest.

\section{Data availability} 
\label{sec_data}
The data that support the findings of this study are openly available in Zenodo at \href{https://zenodo.org/records/11125848}{Seatizen Atlas}.
All code for data processing associated with the current submission is available on \href{https://github.com/SeatizenDOI/drone-upscaling}{drone-upscaling Github}.

The code for downloading data associated with the current submission is available on \href{https://github.com/SeatizenDOI/zenodo-tools}{zenodo-tools Github}.

The code used to train the neural network model used in the current submission is available on \href{https://github.com/SeatizenDOI/DinoVdeau}{DinoVdeau Github}.
\section{Declaration of Generative AI and AI-assisted technologies in the writing process}
During the preparation of this work the author(s) used ChatGPT-4o and GitHub Copilot in order to improve language and readability.
After using this tool/service, the author(s) reviewed and edited the content as needed and take(s) full responsibility for the content of the published article.
This tools were not involved in the design, implementation, data analysis, or manuscript preparation of the study.

\bibliographystyle{unsrtnat}
\bibliography{drone_upscaling}

\appendix
\section{ASV supplementary informations}
\label{annex:asv}
\subsection{Time synchronization}
\label{subsubsec_time_sync}
Videos were cut into frames with a rate of \(2.997 \, \text{fps}\), so that the cutting frame rate \( f_c = 2.997 \, \text{fps} \) is a divisor of the video frame rate \( f_v = 23.976 \), ensuring that the ratio \( \frac{f_v}{f_c} \) is an integer.

Since we use time in order to synchronize metadata and images, we need a method to assign a precise timestamp to each frame.
Before each data acquisition, as differences of several seconds/minutes can be observed between the clocks of the different devices (the GPS receiver clock is not the same as the camera), the user films the time given by a GPS application on his mobile phone with the camera in order to associate the exact satellite time (UTC+0) to a specific frame or image. 
In the case where the time filmed with the camera follows UTC standards, leap seconds caused by the difference between UTC time and GPS time must be taken into account when synchronizing the GPS position with the images. 
This specific frame can then be used as a starting point to correct the timestamp of all images by using the frame rate \( f_c \) and the number of frames between the starting frame and the frame of interest.

Cutting frames with a rate that is a divisor of the video frame rate is particularly important when working with precise position accuracy. 

\begin{figure*}[ht]
    \centering
    \includegraphics[width=\textwidth]{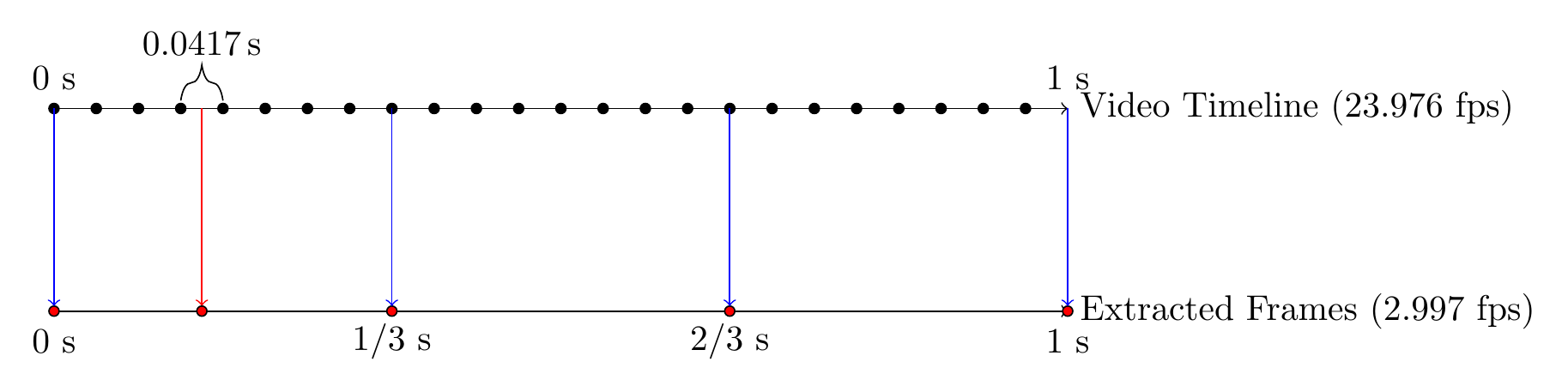}
    \caption{Frame rate extraction from GoPro video. 
    If the cutting frame rate is not a divisor of the video frame rate, a misalignment between the timestamp and the corresponding frame is introduced.
    Later causing an error in assigning the GPS position to the frame.}
    \label{fig:frame_rate_sync}
\end{figure*}

Indeed if the cutting frame rate is not a divisor of the video frame rate, it may happen that during the cutting process of the video a frame is skipped or duplicated, causing a misalignment between the timestamp and the corresponding frame. 
This is represented in Figure \ref{fig:frame_rate_sync}, where blue arrows represent frames extracted from the video timeline with a rate of \( f_c = 2.997 \, \text{fps} \) and red arrow represent a frame extracted with a random frame rate. 
In the first case, there is a perfect alignment between video frames and frames extracted from it. 
In the second case, we can see that for the required frame rate there is no corresponding frame in the original video timeline. 
So that, depending on the chosen option, either the frame is skipped or the closest frame is duplicated.
In both cases, an error is introduced in the extracted frame timestamp.
A difference in the timestamp will result in a misalignment between the real GPS position and the calculated one, which is proportional to the speed at which the ASV acquired the data.
For more information about camera video setting, please refer to Appendix \ref{subsubsec_camera_settings}.

\subsection{Metadata correction}
\label{subsubsec_metadata_corr}
Since the ASV is equipped with a differential GPS, PPK (Post-Processed Kinematic) corrections can be applied to the GPS position of the rover in order to get a centimetric accuracy.

Indeed, for each data collection event a mobile base station has been strategically deployed near the field mission. 
The mobile base station, connected to the \textit{CentipedeRTK} network which provides real-time corrections to the GPS base station, ensures a high precision of the GPS base position \cite{ancelin:hal-04144737}. 
This allows to refine the GPS position of the rover in a post processing step using corrections from the base station.


We can then attach to each frame the corresponding position, using the timestamp as a reference.

\begin{figure}[ht]
    \centering
    \begin{subfigure}[b]{0.4\textwidth}
        \includegraphics[width=\textwidth]{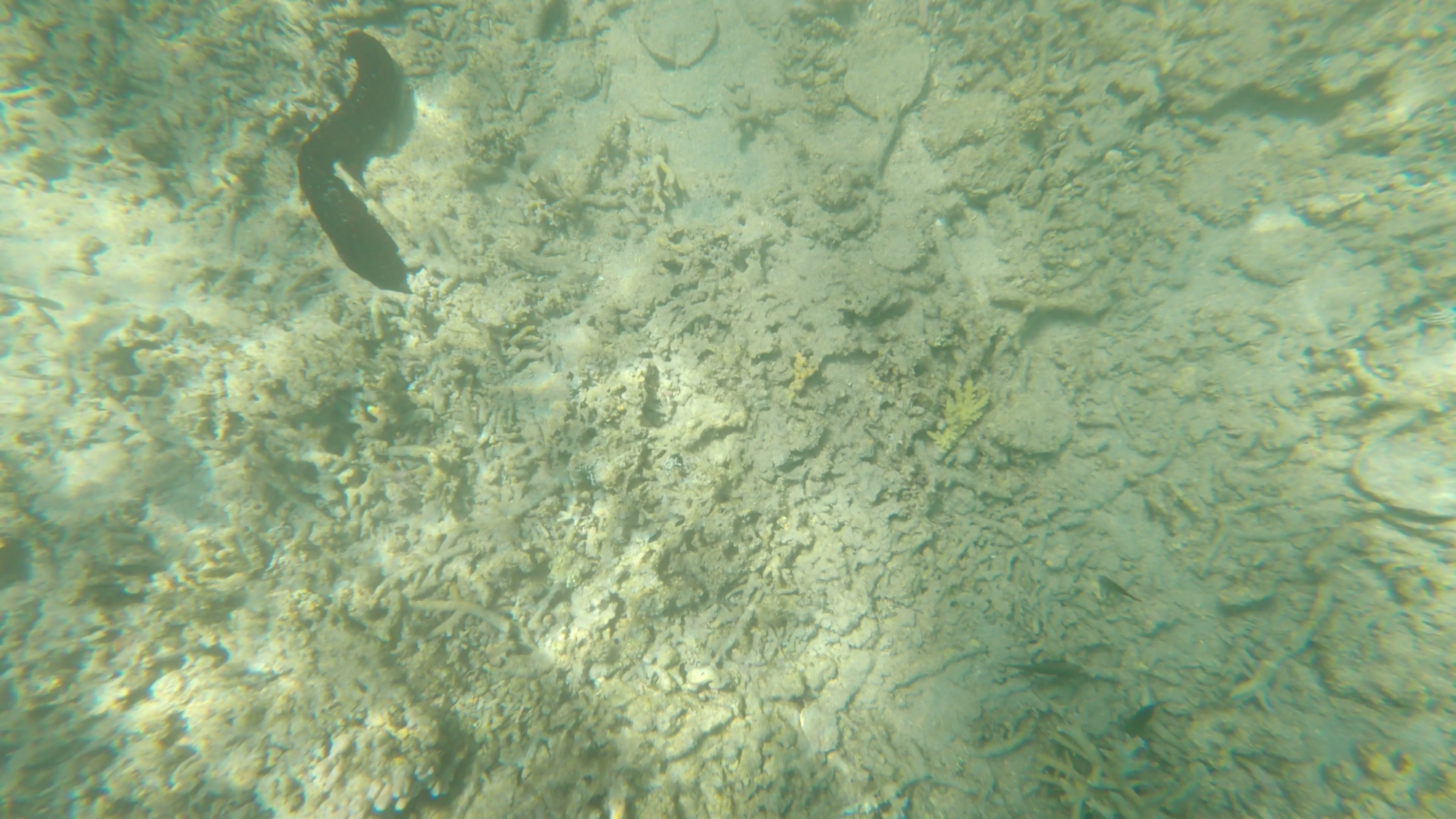}
    \end{subfigure}
    \hspace{2cm}
    \begin{subfigure}[b]{0.4\textwidth}
        \includegraphics[width=\textwidth]{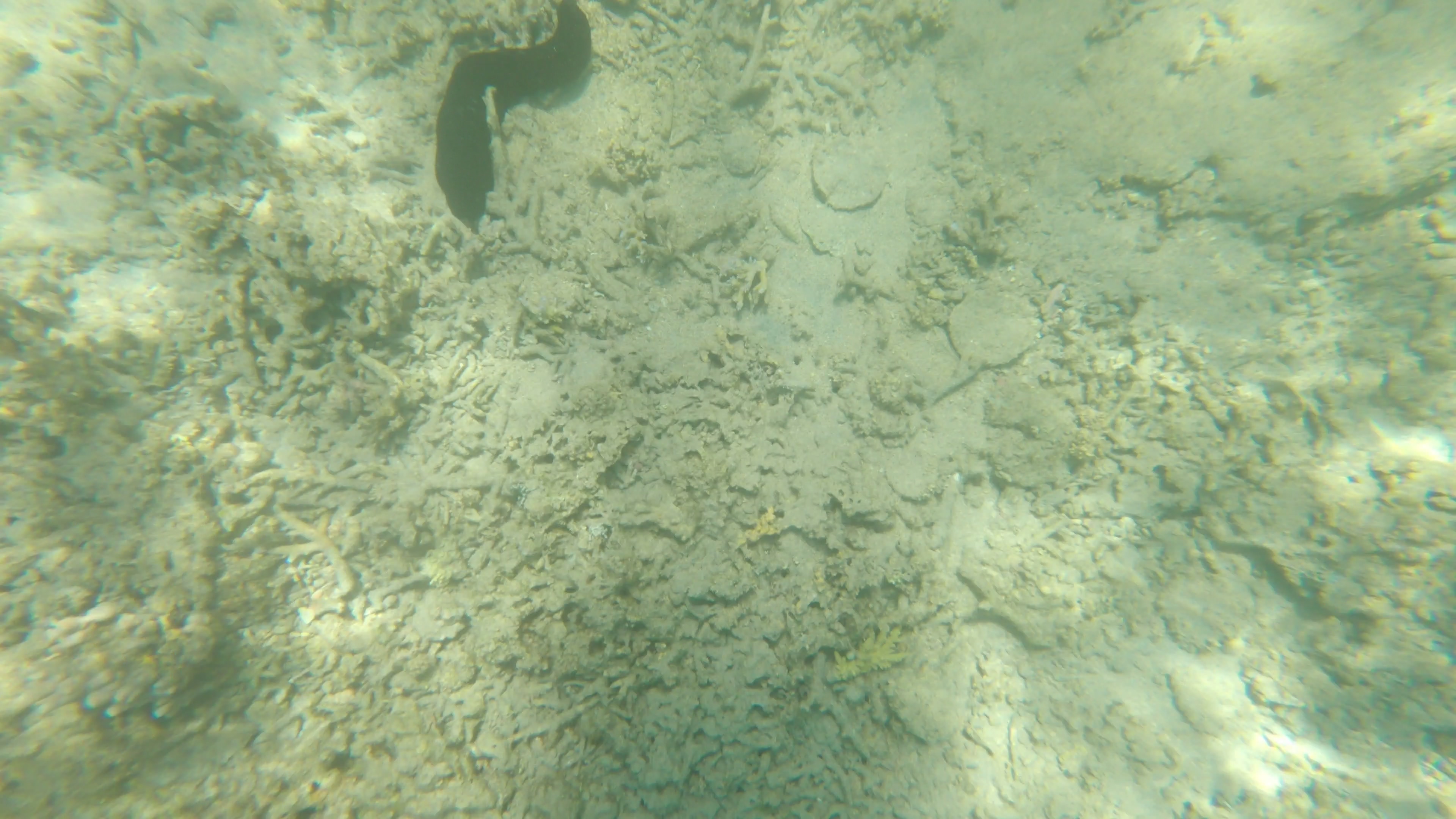}
    \end{subfigure}
    \caption{Example of two images taken in the \textit{Trou d'eau} lagoon in Reunion Island at a distance of 6.062 cm from each other and 24 minutes and 49 seconds apart. The similarity between the two images is an important visual criterion to validate a data collection event.}
    \label{fig:two_images}
\end{figure}

Underwater images positioning was checked in two different ways:
\begin{enumerate}
    \item Firstly we computed the standard deviation on the east and north axis of the GPS position of the rover. 
    If the value was below the centimeter for both axis then the session was considered as a good one. 
    \item Secondly a visual check was done by visually comparing images that had a very close GPS position but a different timestamp. 
    If the two images represent the same zone then the session was considered as a good one.
    An example is given in Figure \ref{fig:two_images}. 
    The two images taken in the \textit{Trou d'eau} lagoon in Reunion Island are at a distance of 6.062 cm from each other and are taken 24 minutes and 49 seconds apart. 
    It is clear how, except from the sea cucumber that has moved a little bit between the two images, the two frames represent the same zone, validating in this way the data collection event.
\end{enumerate}

Moreover, since the ASV is equipped with an IMU (Inertial Measurement Unit) that provides the roll, pitch and yaw angles of the rover, it could be possible to correct the bathymetry data using local geoid parameters and the attitude data of the ASV.
This data were then attached to each frame using, again, the timestamp as a reference.

\subsection{Camera settings}
\label{subsubsec_camera_settings}
GoPro camera setting can be found in Table \ref{tab:video_params}.

\begin{table}[ht]
\centering
\resizebox{0.3\textwidth}{!}{
\begin{tabular}{>{\bfseries}l l}
\toprule
Parameter & Value \\
\midrule
File Type Extension & MP4 \\
MIME Type & video/mp4 \\
Time Scale & 60000 \\
Preferred Rate & 1 \\
Preferred Volume & 1 \\
Firmware Version & HD8.01.02.51.00 \\
Camera Model Name & HERO8 Black \\
Auto Rotation & U \\
Digital Zoom & Y \\
Pro Tune & Y \\
White Balance & 6500K \\
Sharpness & HIGH \\
Color Mode & FLAT \\
Auto ISO Max & 400 \\
Auto ISO Min & 100 \\
Rate & 2\_1SEC \\
Field Of View & L \\
Sensor Readout Time & 7.9200005531311 \\
Electronic Image Stabilization & N/A \\
Image Width & 1920 \\
Image Height & 1080 \\
Graphics Mode & 0 \\
X Resolution & 72 \\
Y Resolution & 72 \\
Compressor Name & GoPro AVC encoder \\
Bit Depth & 24 \\
Video Frame Rate & 59.9400599400599 \\
Avg Bitrate & 45266194 \\
\bottomrule
\end{tabular}
}
\caption{Video File Parameters}
\label{tab:video_params}
\end{table}

\section{UAV supplementary informations}
\label{annex:uav}

\subsection{Mission planning}
\label{susubsec_mission_planning}
\begin{figure*}[ht]
    \centering
    \includegraphics[width=0.5\textwidth]{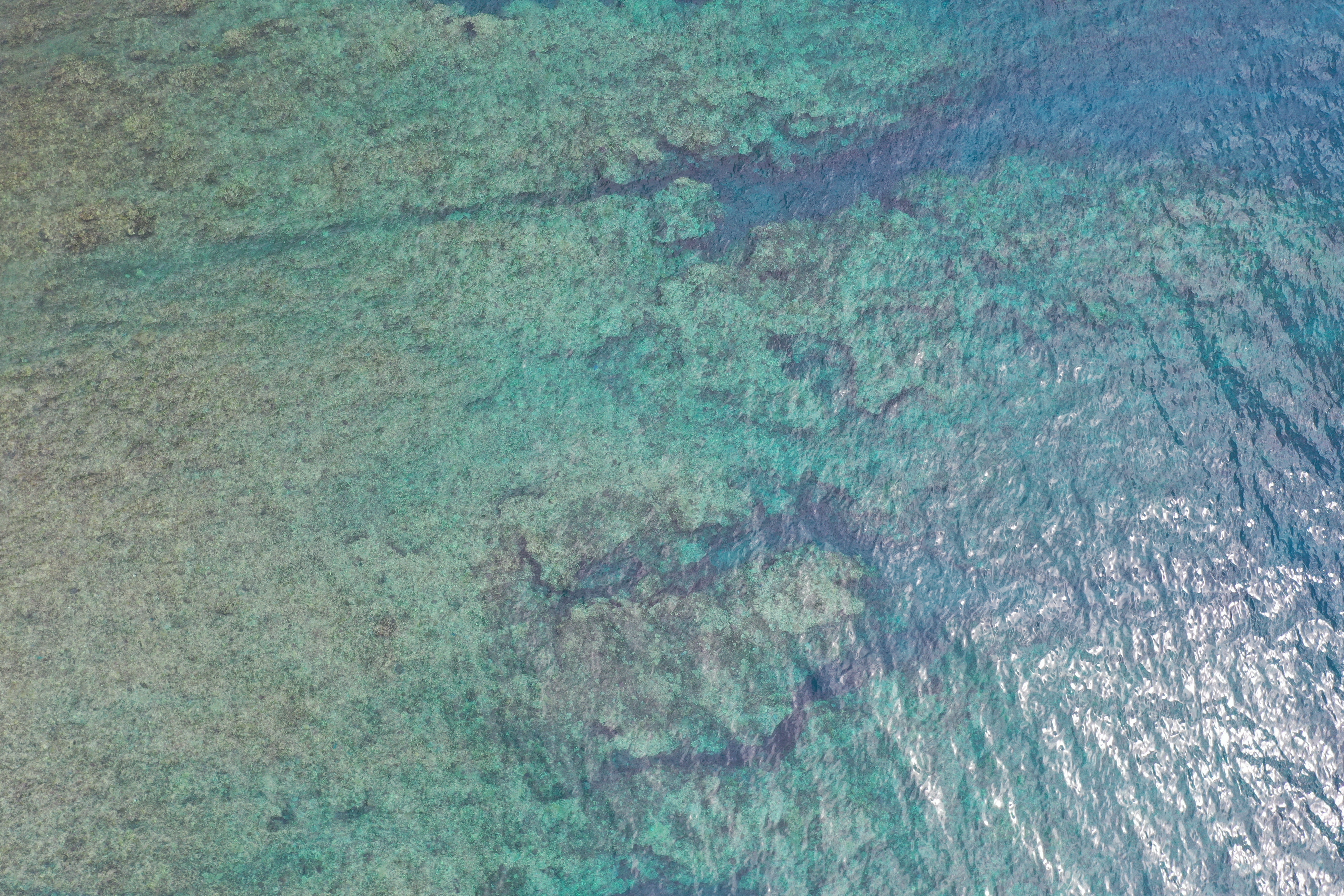}
    \caption{Example of an overexposed image, taken by a Mavic 2 Pro drone in the \textit{Saint-Leu} lagoon in Reunion Island. The sun reflection on the water surface makes it impossible to see the seafloor structure.}
    \label{fig:sun_reflection}
\end{figure*}

Imaging the seabed, even in tropical environments with very clear waters, is often complicated by reflections of sunlight from the water surface.
 Direct sun rays reflections can be extremely bright and cause oversaturated areas on aerial images. 
 These reflections overexpose the image, making it difficult to see the seafloor structure, as shown in Figure \ref{fig:sun_reflection}. 
 
 If the seafloor is not visible in the image, SfM algorithms cannot map these areas effectively. 
 It is thus necessary to avoid these reflections in the images. 
 A solution is to take images when the sun is at a low angle, which reduces the reflection of sunlight on the water surface \cite{slocum2019guidelines}. 
 This can be achieved by surveying the desired area early in the morning or late in the afternoon, or on a cloudy day. 
 In our case, images were taken between daybreak and sunrise, which is the best time to avoid reflections on the water surface on the west part of Reunion island. 

\subsection{Mission execution}
\label{susubsec_mission_execution}
In SfM surveys, overlap and sidelap refer to the percentage to which each consecutive image overlaps with the preceding image and the images on adjacent flight lines. 
In order to obtain a good 3D model, it is important to have a high overlap and sidelap between images. 
The overlap and sidelap should be at least 70\% for a good 3D model \cite{WESTOBY2012300}. 
In our case, the overlap and sidelap were both set to 80\%.

For what concerns the flight altitude, a lower one implies a higher resolution point cloud in the SfM process since more details of the seabed will be visible but on the other hand the wave induced refraction is more visible. 
On the contrary, setting a too high flying altitude will reduce the resolution of the point cloud, making it difficult to distinguish characteristics of study objects (e.g., coral colonies, habitats, etc.). 
In our case, the flight altitude was set to 60 meters.

\subsection{Image processing}
\label{susubsec_image_processing}
Many photogrammetry softwares are available to build a 3D model from images. 
In our case, since the objective was to create an open-source pipeline, the software package \texttt{OpenDroneMap} was used. 
\texttt{OpenDroneMap} is a commercial-grade open-source software package for SfM photogrammetric processing (initially developed for aerial images) that can be used to generate georeferenced orthophotos, point clouds, elevation models and textured 3D models from aerial images.

Settings used for the processing images from both missions are shown in Table \ref{table:odm-settings}.

\begin{table}[ht]
    \centering
    \begin{tabular}{|l|l|}
    \hline
    \textbf{Setting} & \textbf{Value} \\ \hline
    Auto-boundary & True \\ \hline
    DEM Resolution & 2.0 \\ \hline
    DSM & True \\ \hline
    Orthophoto Resolution & 1.0 \\ \hline
    Point Cloud Quality & Ultra \\ \hline
    Rolling Shutter & True \\ \hline
    \end{tabular}
    \caption{OpenDroneMap Processing Settings}
    \label{table:odm-settings}
\end{table}

\subsection{Orthophoto georeferencing}
\label{susubsec_orthophoto_georeferencing}
\begin{figure}[ht]
    \centering
    \begin{subfigure}[b]{0.4\textwidth}
        \includegraphics[width=\textwidth]{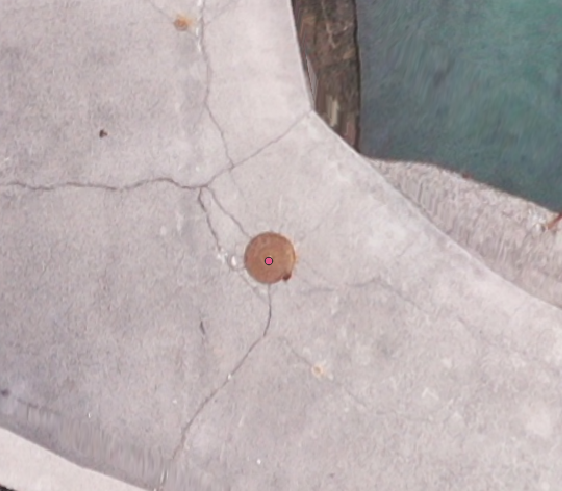}
    \end{subfigure}
    \hspace{2cm}
    \begin{subfigure}[b]{0.4\textwidth}
        \includegraphics[width=\textwidth]{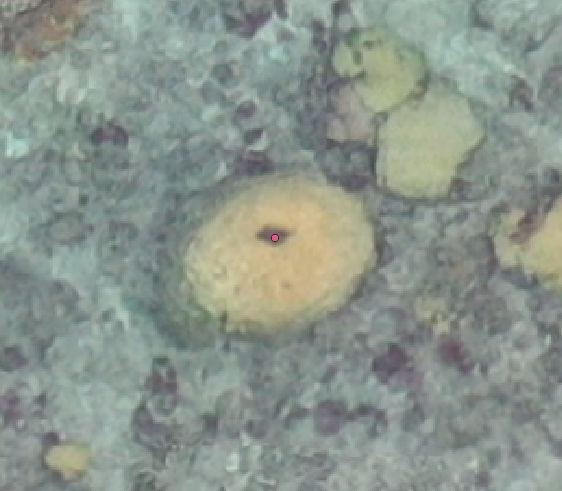}
    \end{subfigure}
    \caption{Examples of Ground Control Points (GCPs) collected in Reunion island in order to correct the aerial orthophoto position.
    Both human made objects (on the left) and easily distinguishable corals (a \textit{Porites} coral on the right) can be used to set GCPs.}
    \label{fig:gcp}
\end{figure}

Since the drone was not equipped with a differential GPS, once the orthophoto was built, Ground Control Points (GCPs) were chosen on fixed and easily distinguishable objects on land (e.g. manhole covers, corners of basketball courts, etc.) and easy-to-distinguish corals (e.g., large \textit{Porites} or $\AcroporeT$ corals). 
GCPs position was then collected using a GPS with centimetric accuracy and then the orthophoto was reconstructed by forcing pixels representing GCPs to be at the ground truth collected position.
GCPs examples are shown in Figure \ref{fig:gcp} with a pink point on the image.

The  \textit{Trou d'eau} lagoon mission measured a surface of 189,682 m\(^2\) and 10 GCP were collected. The \textit{Saint-Leu} lagoon mission measured a surface 204,748 m\(^2\) and 9 GCP were collected.

Final orthophoto positioning was checked by computing the Root Mean Square Error (RMSE) between the GCPs and the orthophoto. 

\begin{table}[ht]
    \centering
    \resizebox{0.5\columnwidth}{!}{
    \begin{tabular}{lcccc}
    \toprule
    \textbf{Lagoon} & \textbf{Error Type} & \textbf{Mean} & \textbf{Standard Deviation} & \textbf{RMS Error} \\
    \midrule
    \textbf{Trou d'eau} & X Error (meters) & 0.001 & 0.011 & 0.011 \\
    & Y Error (meters) & -0.001 & 0.010 & 0.010 \\
    & Z Error (meters) & -0.011 & 0.028 & 0.030 \\
    & \textbf{Total} & & & \textbf{0.022} \\
    \midrule
    \textbf{Saint-Leu} & X Error (meters) & -0.001 & 0.003 & 0.003 \\
    & Y Error (meters) & -0.000 & 0.001 & 0.001 \\
    & Z Error (meters) & 0.000 & 0.004 & 0.004 \\
    & \textbf{Total} & & & \textbf{0.003} \\
    \bottomrule
    \end{tabular}
    }
    \caption{GCP errors statistics for  \textit{Trou d'eau} Lagoon and \textit{Saint-Leu} Lagoon}
    \label{tab:gcp_errors}
\end{table}

In Table \ref{tab:gcp_errors} we show the error statistics for GCPs collected in the \textit{Trou d'eau} and the \textit{Saint-Leu} lagoons. 
We can observe that the RMSE is below 2.5 cm for both lagoons, which is a satisfactory level of precision for the continuation of the study.

\section{Fine scale deep-learning model} 
\label{sec_finescal_deepmodel}

\textit{DinoV2}, a family of state-of-the-art transformer models in computer vision produces general-purpose visual features (i.e., features that work across image distributions and tasks without fine-tuning), and is compatible with classifiers as simple as linear layers.
Meaning that the model can be readily applied to various tasks like image classification or segmentation without necessitating encoder fine-tuning \cite{oquab_dinov2_2023}.
This implies that instead of training the entire model on a specific dataset, we can simply fine-tune a classification head atop the frozen encoder.\footnote{To give an idea, the large version of the model (\texttt{Vit-L}) has 0.3B of parameters and the giant version (\texttt{Vit-g}) 1.1B parameters.}\textit{DinoV2} comprises four distinct models based on different sizes: small, base, large, and giant. 

In our study, following a performance evaluation against training time experiment, we opted for the large model.
Transitioning from the small to base version and from base to large version, demonstrates a performance increase for all the metrics (F1 Micro, F1 Macro, Roc Auc, and Accuracy) with a reasonable uptick in training time, see Table \ref{tab:model_performance}. 
 The giant version of the model, despite having slightly lower loss than the large model, has lower performances for all the metrics and the training steps per second are almost multiplied by three, inducing a considerable increase in total training time.

\begin{table}[ht]
\centering
\resizebox{0.5\columnwidth}{!}{
\begin{tabular}{lcccccc}
\toprule
Model & Loss & F1 Micro & F1 Macro & Roc Auc & Accuracy & Steps per second \\
\midrule
DinoVdeau-small & 0.1320 & 0.8009 & 0.6614 & 0.8649 & 0.2903 & \textbf{0.164} \\
DinoVdeau-base  & 0.1260 & 0.8131 & 0.6976 & 0.8760 & 0.3014 & 0.207 \\
DinoVdeau-large & 0.1209 & \textbf{0.8228} & \textbf{0.7175} & \textbf{0.8813} & \textbf{0.3111} & 0.21 \\
DinoVdeau-giant & \textbf{0.1208} & 0.8209 & 0.7101 & 0.8812 & 0.3080 & 0.66 \\
\bottomrule
\end{tabular}
}
\caption{Model size comparison}
\label{tab:model_performance}
\end{table}

\begin{figure*}[ht]
\centering
\includegraphics[width=\textwidth]{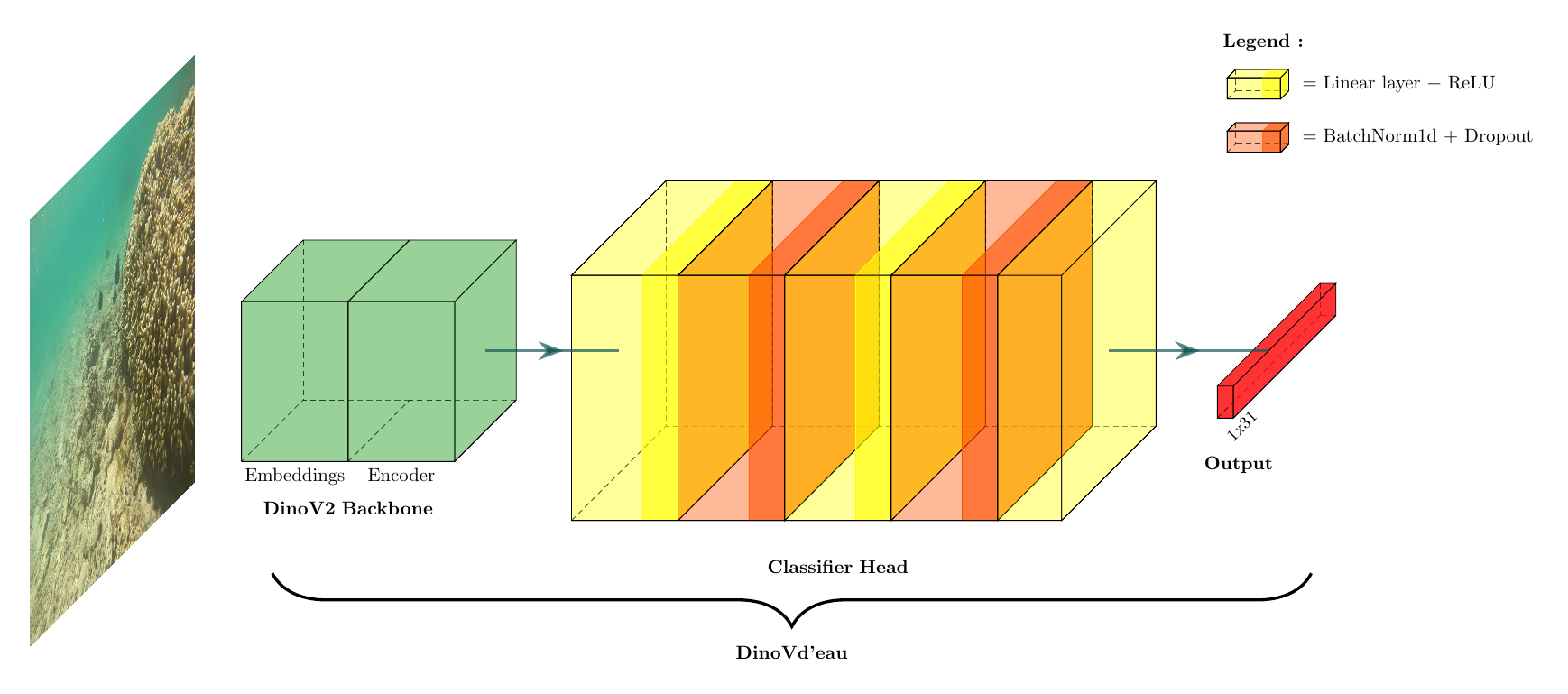}
\caption{Architecture of \textit{DinoVdeau} model}
\label{fig_DinoVd_eau_architecture}
\end{figure*}

For the classification head, we used a more complex model than a simple linear classifier in order to improve the expressiveness of the model and capture more complex interactions between the variables extracted by the backbone model. 
Specifically, we added a bottleneck layer block consisting of a linear layer, a ReLU, a batch normalization, and a dropout layer. The resulting model is called \textit{DinoVdeau} and the architecture is shown in Figure \ref{fig_DinoVd_eau_architecture}. 

Introducing the bottleneck layer block instead of a single linear classifier results in a performance gain of approximately $0.006\%$ and $0.015\%$ for the F1 Micro and the Accuracy metrics respectively.

Using \textit{DinoV2} as the backbone compared to a \textit{Resnet50} baseline model results in a performance gain of approximately $0.098\%$ and $0.140\%$ for the F1 Micro and the Accuracy metrics respectively.

The model was trained for 93 epochs, incorporating an early stopping mechanism with a patience of 10 epochs to mitigate overfitting. 
Maintaining a fixed batch size of 32, the initial learning rate was established at $10^{-3}$, and was decreased by a factor of $0.1$ whenever the model's performance plateaued for more than 5 epochs. 
Network weight updates were executed using the Adam optimization algorithm with a weight decay of $10^{-4}$.

The training lasted 75 hours, on the \textit{Datarmor} supercomputer equipped with an \texttt{NVIDIA Tesla V100 PCIe 32 GB} GPU and \texttt{32 Intel Xeon-Gold 4216 (2.1GHz/16-core/100W)} CPUs. 
All the trainings were done using the Huggingface \texttt{Transformers} library \cite{wolf-etal-2020-transformers}. 

\section{Data splitting} 
\label{sec_data_splitting}
As explained in \cite{10.1093/bioinformatics/btz421}, the process of data splitting is a crucial step in the construction of image classification models. 
This is essential to assess their effectiveness with an unbiased approach.
To achieve this, we employ preprocessing techniques and leverage prior knowledge of the dataset. 
The approach involves dividing the samples into three distinct parts: the training set $D_{tr}$, the validation set $D_{val}$, and the test set $D_{test}$. 
It's important to note that our task involves multilabel classification, where an input can have multiple labels. 
This complicates the stratification process compared to monolabel classification. 
Traditional single-label approaches to stratifying data fall short in providing balanced dataset divisions in the multilabel case because it's not feasible to create tuples (image, label), given that each image corresponds to a variable number of labels. 
Consequently, once an image is assigned to a dataset, all corresponding labels are assigned to it.

To build a model with high generalization capabilities, we opt for a temporal criterion to independently divide the dataset into three subdatasets. 
The goal is to maximize the diversity of the training dataset by including images from different periods and islands in the Indian Ocean. 
This temporal splitting criterion corresponds to a spatial one as well, considering that images are typically collected during data collection campaigns in specific locations. 
This approach ensures that the training dataset is not only temporally diverse but also representative of various spatial contexts.
Subsequently, we employ the \texttt{scikit multilabel} data stratification technique to split each subdataset into three subsets \cite{10.1007/978-3-642-23808-6_10}. 
For the purpose of achieving optimal predictive performance for our neural network, we implement the algorithm represented by the following pseudo-code:

\begin{algorithm}[H]
	\caption{Dataset splitting algorithm}
	\label{split_algo1}
	\begin{algorithmic}[1]
		\For{$i = 1,\dots, Nb\_years$}
		\State 1.Split the $D_{i}$ into \textit{$train_i$} and \textit{$val-test_i$} sets using the Multi-label data stratification technique
		\State 2.Split the \textit{$val-test_i$} set into \textit{$val_i$} and \textit{$test_i$} sets using the Multi-label data stratification technique.
		\State 3.Concatenate the current \textit{$train_i$}, \textit{$val_i$}, and \textit{$test_i$} datasets to the overall $D_{train}$, $D_{val}$ and $D_{test}$datasets.
		\EndFor
	\end{algorithmic}
\end{algorithm}

In Table \ref{tab_class_freq}, we present the total number of images for each class, along with their corresponding distribution in the training, validation, and test sets. 
The results indicate a well-balanced class distribution.

\begin{table*}[ht]
\centering
\resizebox{0.6\textwidth}{!}{
\begin{tabular}{|l|c|c|c|c|}
\hline
\textbf{Class} & \textbf{Train Frequency} & \textbf{Validation Frequency} & \textbf{Test Frequency} & \textbf{Total} \\
\hline
Acropore\_branched & 0.666666 & 0.167702 & 0.165632 & 1206 \\
\hline
Acropore\_digitised & 0.690607 & 0.160059 & 0.149552 & 674 \\
\hline
Acropore\_tabular & 0.640284 & 0.183073 & 0.177643 & 1507 \\
\hline
Algae\_assembly & 0.609174 & 0.194251 & 0.195574 & 3562 \\
\hline
Algae\_limestone & 0.601715 & 0.198876 & 0.199408 & 2207 \\
\hline
Algae\_sodding & 0.606493 & 0.197214 & 0.196293 & 3426 \\
\hline
Dead\_coral & 0.613217 & 0.194557 & 0.193226 & 1839 \\
\hline
Fish & 0.643513 & 0.169074 & 0.169413 & 1359 \\
\hline
Human\_object & 0.599117 & 0.198823 & 0.200059 & 678 \\
\hline
Living\_coral & 0.604631 & 0.198982 & 0.195387 & 2916 \\
\hline
Millepore & 0.612976 & 0.208060 & 0.178964 & 571 \\
\hline
No\_acropore\_encrusting & 0.602345 & 0.207921 & 0.188734 & 682 \\
\hline
No\_acropore\_foliaceous & 0.742105 & 0.119298 & 0.136842 & 285 \\
\hline
No\_acropore\_massive & 0.595306 & 0.204966 & 0.200388 & 1548 \\
\hline
No\_acropore\_sub\_massive & 0.623316 & 0.187305 & 0.188379 & 1930 \\
\hline
Rock & 0.606011 & 0.197692 & 0.196297 & 6171 \\
\hline
Sand & 0.599165 & 0.200334 & 0.199501 & 5990 \\
\hline
Scrap & 0.590874 & 0.201673 & 0.207453 & 3586 \\
\hline
Sea\_cucumber & 0.600769 & 0.195385 & 0.204615 & 1300 \\
\hline
Sea\_urchins & 0.589441 & 0.186291 & 0.224267 & 321 \\
\hline
Sponge & 0.581487 & 0.192027 & 0.226486 & 389 \\
\hline
Syringodium\_isoetifolium & 0.600307 & 0.198256 & 0.201437 & 1949 \\
\hline
Thalassodendron\_ciliatum & 0.600921 & 0.199232 & 0.199846 & 1304 \\
\hline
Useless & 0.601836 & 0.199179 & 0.199985 & 977 \\
\hline
\end{tabular}}
\caption{Class frequency distribution across training, validation, and test sets.}
\label{tab_class_freq}
\end{table*}

\end{document}